\documentclass[11pt, a4paper, gdm]{google}

\usepackage{amsmath}
\usepackage{amssymb}
\usepackage{enumitem}
\usepackage{parskip} % Adds spacing between paragraphs
\usepackage{algorithm}
\usepackage{algpseudocode}
\usepackage{wrapfig}
\usepackage{subcaption}
\usepackage[most]{tcolorbox}
\usepackage{booktabs}
\usepackage{tabularx}
\usepackage{multirow}

\usepackage[utf8]{inputenc}
\usepackage{geometry}
\usepackage{xcolor}
\usepackage{tcolorbox}
\tcbuselibrary{skins, breakable}

\newtcolorbox{promptbox}[2][]{%
    colback=gray!5!white,      
    colframe=gray!60!black,    
    coltitle=white,            
    title={#2},                
    fonttitle=\bfseries\small,
    fontupper=\footnotesize,
    breakable,                 
    enhanced,                  
    boxrule=0.8pt,             
    left=4pt, right=4pt, top=4pt, bottom=4pt,
    parbox=false,
    #1                         
}

\newcommand{\code}[1]{\texttt{\textcolor{blue!60!black}{#1}}}

\usepackage[authoryear, sort&compress, round]{natbib}

\definecolor{aszlam}{RGB}{120, 0, 140} % Example value
\definecolor{ldery}{RGB}{0, 120, 88}
\definecolor{ranzato}{RGB}{0, 70, 140} % Example value
\definecolor{myLightBlue}{RGB}{150, 200, 250} % Example values
\definecolor{mr}{RGB}{170,100,100}
\definecolor{qixuan}{RGB}{77,136,240}
\definecolor{jj}{RGB}{115,201,122}
\usepackage{multirow}

\uselogo{} 

\title{
Context Training with Active Information Seeking
}

\correspondingauthor{zeyu.huang@ed.ac.uk,akuncoro@google.com}

\reportnumber{0001} % Leave blank if n/a

\renewcommand{\today}

\author[1,2]{Zeyu Huang}
\author[2]{Adhiguna Kuncoro}
\author[2]{Qixuan Feng}
\author[2]{Jiajun Shen}
\author[2]{Lucio Dery}
\author[2]{Arthur Szlam}
\author[2]{Marc'Aurelio Ranzato}

\affil[1]{The University of Edinburgh}
\affil[2]{\thepa{}{}}

\begin{abstract}
Most existing large language models (LLMs) are expensive to adapt after deployment, especially when a task requires newly produced information or niche domain knowledge. Recent work has shown that, by manipulating and optimizing their context, LLMs can be tailored to downstream tasks without updating their weights.
However, most existing methods remain closed-loop, relying solely on the model's intrinsic knowledge. 
In this paper, we equip these context optimizers with Wikipedia search and browser tools for active information seeking. We show that naively adding these tools to a standard sequential context optimization pipeline can actually degrade performance compared to baselines. However, when paired with a search-based training procedure that maintains and prunes multiple candidate contexts, active information seeking delivers consistent and substantial gains. We demonstrate these improvements across diverse domains, including low-resource translation (Flores+), health scenarios (HealthBench), and reasoning-heavy tasks (LiveCodeBench and Humanity’s Last Exam). Furthermore, our method proves to be data-efficient, robust across different hyperparameters, and capable of generating effective textual contexts that generalize well across different models. 
\end{abstract}

\begin{document}

\maketitle

\section{Introduction}
The rise of Large Language Models (LLMs)~\citep{DBLP:journals/corr/abs-2507-06261,singh2025openai} represents a fundamental shift away from task-specific AI. Unlike their predecessors that were trained on narrow domains, contemporary LLMs exhibit impressive general-purpose capabilities~\citep{DBLP:conf/coling/YadavB18,DBLP:journals/widm/ZhangWL18}, allowing them to navigate diverse domains and scenarios~\citep{DBLP:journals/corr/abs-2502-06807,DBLP:journals/corr/abs-2505-23281,li-etal-2025-investorbench,DBLP:journals/corr/abs-2503-24047,DBLP:journals/corr/abs-2507-01679,DBLP:conf/iclr/HuangQWPT25}. 
Yet, once deployed, these models remain difficult to adapt continuously when a task requires newly produced information, niche domain knowledge, or behavior specialized to unfamiliar settings~\citep{DBLP:journals/corr/abs-2507-21046,DBLP:journals/corr/abs-2508-07407}.
Retraining or fine-tuning models with new data is a plausible solution, but it incurs prohibitive training costs and risks catastrophic forgetting. Consequently, several works have proposed shifting the focus \textit{from updating model parameters} to \textit{optimizing the model context}, \textit{i.e.}, constructing an evolving working memory to adapt to new tasks.~\citep{DBLP:journals/tmlr/WangX0MXZFA24,cheng2024trace, liu-etal-2025-contextual}. 

Under this paradigm, learning is reformulated as the iterative refinement of the input context or a pluggable memory bank, rather than an update to the parameters. 
At each iteration, an LLM-based optimizer reflects on a data batch, such as past task attempts and feedback, and then refines the existing context. By selecting, abstracting, and refactoring experiences into a dynamic knowledge base or skill set, such systems can achieve positive transfer on relevant tasks without altering a single parameter~\citep{li2026just}. Pioneering frameworks such as ProTeGi~\citep{pryzant-etal-2023-automatic}, TextGrad~\citep{DBLP:journals/corr/abs-2406-07496}, and DSPy~\citep{khattab2024dspy} demonstrate the promise of this approach for reasoning and code-generation tasks without manual prompt engineering.
%Throughout the paper, we use \textit{agent} to denote an LLM invocation with a role-specific instruction and, when applicable, tool access.

Despite their initial success, most existing approaches are constrained by a fundamental drawback: They are closed systems. 
Lacking external grounding and access to external sources of information, these frameworks primarily rearrange and refine the optimizer's existing internal knowledge, making it difficult to incorporate task-relevant information that falls outside the model's parametric memory. 
This creates a critical bottleneck: 
When the desirable information (\textit{e.g.}, a report released after training or a niche technical fact) lies outside the model's frozen parametric knowledge, the optimizer cannot reliably discover it and make effective updates to the context. 
This issue is especially acute when the training feedback identifies that the executor is wrong, but does not itself contain the missing knowledge needed to repair the context. In that case, a closed optimizer can only reorganize or extrapolate from its existing knowledge, and the resulting update is written directly into the context used by future executor calls. Moreover, the system may amplify hallucinations rather than verify the ground truth. 
As noted in recent studies on the ``curse of recursion''~\citep{shumailov2024ai}, such self-consuming loops without external data could lead to \textit{context collapse}~\citep{DBLP:journals/corr/abs-2510-04618}, where the diversity and utility of the optimized context suddenly degrade.

Nevertheless, simply granting the optimizer access to the web does not guarantee success. We identify two critical failure modes in the standard sequential training pipeline:
(1) \textbf{Context Pollution}: Given the uncontrolled nature of web content, the model risks injecting low-quality or misleading information into the context. According to our preliminary study, the optimizer agents struggle to recover from these context-polluting updates, especially when the context optimizer agent lacks an explicit backtracking mechanism.
(2) \textbf{Local Optima}: During training, a greedy optimization strategy may commit to sub-optimal trajectories early on, achieving only marginal gains while missing better-performing alternative solutions. 

To address these two issues, we adopt a beam-search-style training process that maintains a pool of candidate contexts, explores diverse updates in parallel, and discards trajectories that are contaminated by low-quality external data or trapped in weak strategies. We also include the current best context in the candidate pool as a ``Do Nothing'' option. This ensures that, if \emph{all} new explorations turn out to be noisy or unhelpful, the optimizer can simply retain the previous best state.
We examine our proposed approach across diverse domains, including low-resource translation (Flores+), healthcare (HealthBench), and
two reasoning-heavy tasks (LiveCodeBench \& Humanity's Last Exam).
We observe consistent performance gains when active information seeking is paired with this search-based training procedure, compared to the sequential, closed-context training baseline, \emph{without} applying any manual task-specific optimization or task-specific prompt tuning.
Moreover, we present analysis and ablation studies showing that the method is data-efficient and robust to different hyperparameters, and that the context optimized for one model generalizes well to other models.

\section{Related Work: from Context Engineering to Working-Memory Evolution} 
\label{sec:related works}

LLMs' in-context learning capability offers a promising way to enhance model performance and elicit desired behaviors without updating model parameters. We survey this progression from \textit{Context Engineering} to \textit{Working-Memory Evolution}; the former focuses on the strategic composition of the model's final input to maximize immediate performance, while the latter attempts to establish a dynamic workspace to enable efficient adaptation to new tasks and environments.

\paragraph{Context Engineering} 
Context engineering encompasses a broad spectrum of techniques designed to optimize the information distribution within the input of a frozen LLM~\citep{mei2025survey}.
Early research established the foundation of this paradigm through few-shot prompting.~\citep{DBLP:conf/nips/BrownMRSKDNSSAA20,DBLP:journals/csur/Song0CMS23,DBLP:conf/emnlp/Dong0DZMLXX0C0S24}. 
Prompt engineering then became popular~\citep{DBLP:journals/corr/abs-2406-06608,DBLP:journals/corr/abs-2402-07927} and evolved along two distinct lines: (1) principled heuristics, such as the widely adopted Chain-of-Thought prompting~\citep{DBLP:conf/nips/Wei0SBIXCLZ22}; (2) automated optimization strategies that utilize the LLM itself to iteratively refine the prompt via genetic algorithms~\citep{DBLP:conf/icml/FernandoBMOR24} and Beam Search~\citep{pryzant-etal-2023-automatic}.
Furthermore, context engineering extends to external augmentation such as Retrieval-Augmented Generation (RAG)~\citep{zhang2025survey,amugongo2025retrieval,li2025retrieval} and tool-use that injects relevant documents or execution outputs into the context window.
Overall, whether through the precise tuning of instructions or the integration of external knowledge pieces, the unified goal is to construct a composite input that maximizes inference capability. Retrieval-Augmented Generation typically assumes an existing corpus or database and focuses on retrieving the right evidence from it for a given query. Our work notably differs from RAG because our optimizer agent actively seeks missing information, constructing and editing the evolving knowledge base from executor feedback, rather than relying solely on a fixed corpus and embedding similarity based retriever. Notably, recent theoretical perspectives suggest that this optimization process functions as a form of pseudo-gradient descent in the discrete token space, navigating the model's landscape without parameter updates~\citep{DBLP:conf/nips/WenJKGGG23,DBLP:journals/corr/abs-2503-20561}.

\paragraph{Self-Evolving Working-Memory} Leveraging the effectiveness of In-Context Learning (ICL) in Large Language Models (LLMs), recent advancements in context engineering aim to transform static context into a dynamic working memory~\citep{DBLP:journals/corr/abs-2310-08560,DBLP:journals/corr/abs-2502-12110}. This evolution facilitates efficient task adaptation~\citep{DBLP:journals/corr/abs-2507-05257,DBLP:journals/corr/abs-2508-16153,DBLP:journals/corr/abs-2510-04618} and online continual learning~\citep{DBLP:conf/icml/WangMFN25,DBLP:journals/corr/abs-2509-25140,liu-etal-2025-contextual,momeni2025context}. While specific implementations vary across domains, most methods can be formulated within a unified dual-component framework: (1) an executor agent for trace collection, and (2) one or more optimizer agents that analyze and abstract these traces into reusable knowledge and skills, which are subsequently consolidated into a memory bank. This framework has demonstrated strong performance in adapting LLM agents to unfamiliar agentic tasks~\citep{DBLP:journals/corr/abs-2510-04618,zhang2026expseek}, games~\citep{DBLP:journals/tmlr/WangX0MXZFA24,he2025evotest,wei2025evo}, and general problem-solving scenarios~\citep{xu2025metatextgrad,DBLP:journals/corr/abs-2504-07952,cai2025flex}. However, the context optimization stage in these methods typically operates as a \emph{closed} system, relying fully on environmental feedback and the internal Reflection capabilities~\citep{shinn2023reflexion,DBLP:conf/nips/MadaanTGHGW0DPY23} of the optimizer agent. This limitation prompts a critical question: What if the optimizer agent lacks the prerequisite knowledge to update the context effectively? Furthermore, could the optimizer agent actively search for information, rather than relying solely on thousands of closed-loop trial-and-error iterations?

To bridge this gap, we study how to equip the optimizer agent with information-seeking capabilities so that it can retrieve external information during context optimization. Unlike prior closed-loop methods, our focus is on whether external grounding can improve the optimizer's context updates when the required knowledge is not already stored in the model.
To keep the study as general as possible, we adopt a simple training framework and context design rather than adding task-specific engineering. This also extends to the prompts in App.~\ref{app:prompts}, which are shared across domains instead of being specially optimized for any one benchmark. Our empirical results show that the standard sequential training pipeline remains constrained by the model's frozen knowledge, whereas external grounding becomes effective when paired with a search-based training procedure. Because this change targets the context optimization stage, it is largely orthogonal to the surrounding agent workflow and can be integrated into many existing approaches and agent harnesses.

\section{Methodology}

\subsection{Preliminary: Learning as State Optimization}
We begin with a general view of learning as state optimization, which places parameter and context training under the same perspective.
Specifically, a \textit{general learning system} minimizes the divergence between the prediction and the desired outcomes with the following components $\Lambda = \langle \mathcal{M}, \mathcal{S}, \mathcal{O}, \mathcal{D}, R \rangle$: 
\begin{itemize}
    \item $\mathcal{M}: \mathcal{X} \times \mathcal{S} \rightarrow \mathcal{Y}$: An inference function mapping inputs $x \in \mathcal{X}$ and state $S \in \mathcal{S}$ to the system's prediction $\hat{y} \in \mathcal{Y}$.
    \item $\mathcal{S}$: The modifiable state space encoding the system's knowledge. In standard deep learning systems, this state is usually the model parameters; in other settings, it may be a soft prompt, cache, memory bank, or input context. 
    \item $\mathcal{O}: \mathcal{S} \times \mathcal{B} \rightarrow \mathcal{S}$ 
    : An optimizer that updates the state based on a learnable batch $\mathcal{B}$.
    \item $\mathcal{D}$: A distribution over the task space $\mathcal{X}$.
    \item $R$: A reward function indicating the discrepancy between the prediction $\hat{y}$ and the ideal output.
\end{itemize}
These components interact within a cyclic learning pipeline. At each learning step $t$, given a batch of input $X_t \sim \mathcal{D}$, the system generates its corresponding prediction $\hat{Y}_t = \mathcal{M}(X_t; S_t)$ and receives feedback $r_t = R(X_t, \hat{Y}_t)$, which forms a learnable batch $\mathcal{B}_t = (X_t, \hat{Y}_t, r_t)$. 
The optimizer $\mathcal{O}$ then updates the current state as $S_{t+1} \leftarrow \mathcal{O}(S_t, \mathcal{B}_t)$ to reduce the discrepancy between the generated outputs and the optimal behavior. 
The final objective of the system is to determine the optimal state $S^*$ that maximizes the feedback function $R$ over $\mathcal{D}$:
\begin{equation}
    S^* = \arg\max_{S \in \mathcal{S}} \mathbb{E}_{x \sim \mathcal{D}} [R(x,\mathcal{M}(x; S))]
\end{equation}

Standard gradient-based learning is a specific instantiation of this process, in which the modifiable state is the parameter vector and the optimizer is defined by gradient-based updates. Other instantiations may optimize continuous prompts or cached states with gradients. In this work, we focus on a frozen-weight instantiation where the modifiable state is a discrete, human-readable context $C$.

\subsection{Context Training as a Frozen-Weight Instantiation}
Under this instantiation, \textit{context training} modifies the model's behavior (prediction $\hat{y}$) without altering its weights $\theta$.
We refer to the LLM-based components in this pipeline as \textit{agents} because they are invoked with role-specific instructions and tool access. 
The \textit{executor agent} solves task instances conditioned on the current context, while the \textit{optimizer agent} reads trajectories and feedback collected from the executor agent and updates the context.
As shown in the Fig.~\ref{fig:ctx_training}, context training involves three steps analogous to the role of optimization in gradient-based learning:
\begin{wrapfigure}{r}{0.55\textwidth}
    \centering
    \includegraphics[width=0.55\textwidth]{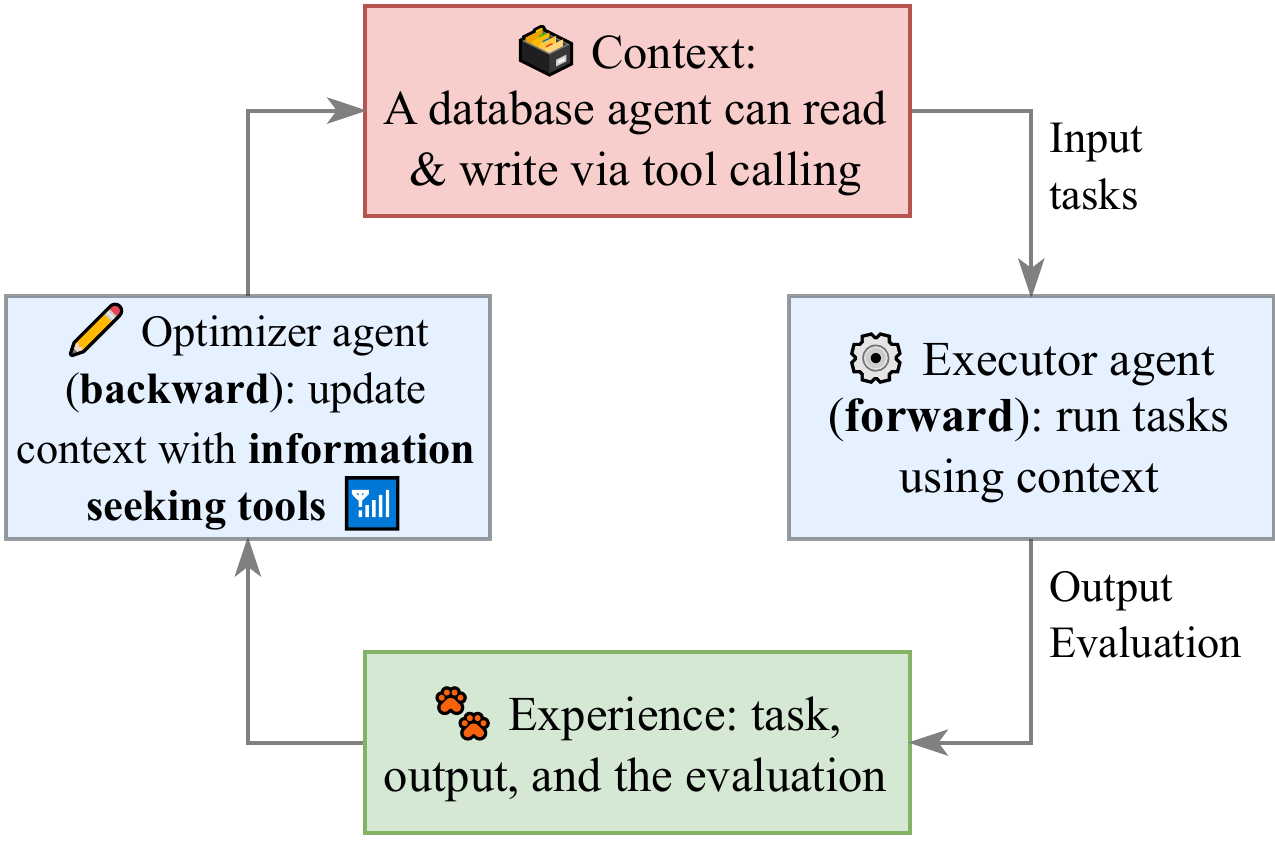}
    \caption{\footnotesize The context training pipeline. In this paper, we propose two main modifications: instantiating the context as a structured database and augmenting the optimizer agent with information-seeking tools.}
    \label{fig:ctx_training}
\vskip -20pt
\end{wrapfigure}
\begin{enumerate}
    \item \textbf{Forward pass:} The executor agent processes the input task conditioned on the existing context.
    
    \item \textbf{Loss function:} The outputs from the executor agent are passed to a reward function to quantify the performance gap. This signal may take the form of a scalar score, a verifiable reward, or natural language feedback diagnosing the error.
    
    \item \textbf{Update step:} An optimizer agent analyzes the feedback and updates the context. In most prior work on context optimization, this step entails rewriting the system prompt to correct errors for the subsequent iteration. Prompts for both agents are detailed in App.~\ref{app:prompts}. In this work, these prompts are kept general-purpose rather than being specially optimized for any particular task.
\end{enumerate}
Furthermore, we introduce two primary modifications, which are presented in detail in the next section:
(1) We instantiate the context as an external structured database that models can read from and write to via function calling;
(2) We augment the optimizer agent with information-seeking tools, enabling it to retrieve missing information from the web, without being limited by its frozen parametric knowledge.

\subsection{Context Management and Information Seeking Tools}

\paragraph{Context Management Tool}
In this work, we instantiate the context as a structured database composed of discrete \textit{resource} items, distinct from the traditional monolithic textual prompt. 
Each \textit{resource} comprises several attributes: (1) a unique resource ID; (2) a concise summary of the item; (3) the raw content; and (4) metadata including the information source, length, keywords, and a text embedding generated by \texttt{gemini-embedding-001}\footnote{\href{https://ai.google.dev/gemini-api/docs/embeddings}{Gemini embeddings documentation}}.
We implement an interface that enables the optimizer agent to interact with this structured database via tool calls. Functionally, the interface supports essential ``write'' operations, including initializing an empty context, and adding, deleting, or updating specific resource items. It also facilitates various ``read'' actions, enabling the model to preview the current context, retrieve specific resources by ID, or search for relevant content via keywords, embeddings, or a dedicated retrieval sub-agent.
Compared to standard monolithic textual prompts, this tool offers greater precision in manipulating context.
It allows the optimizer agent to surgically update or remove specific content without regenerating or reprocessing the entire context, while enabling the executor to retrieve only the resources most relevant to the current task.
A more detailed description of this tool is provided in Tab.~\ref{tab:tool_actions} and App.~\ref{app:tool_actions}.

\paragraph{Information Seeking Tools}
To transcend the closed nature of existing context training pipelines, we equip the agent with external grounding capabilities via two specialized tools:
(1) \textbf{WikipediaSearchTool}: implemented based on the Python \texttt{wikipedia} library\footnote{\href{https://pypi.org/project/wikipedia/}{Python \texttt{wikipedia} package}}, the tool makes it easy to access and parse data from Wikipedia.
(2) \textbf{BrowserUseTool}: this tool enables the agent to navigate web pages dynamically. It can parse HTML content to extract code snippets, recent reports, or documentation that Wikipedia has not yet indexed. This tool is particularly beneficial when the model possesses only vague notions of the desired information. Our implementation leverages the \texttt{browser-use} library\footnote{\href{https://github.com/browser-use/browser-use}{\texttt{browser-use} repository}}.
We include the WikipediaSearchTool to allow the optimizer agent to query specific concepts easily. It is primarily triggered when the optimizer detects declarative knowledge gaps (\textit{e.g.}, missing definitions).
For more complex information-seeking scenarios, we prompt the model to use browsers, as this is a more general way for agents to retrieve information from the web.
By integrating these tools, the optimizer $\mathcal{O}$ transitions from a pure reasoning engine to an active searcher. In our pipeline, before proposing an update $S_{t+1}$, the optimizer can invoke these tools to verify its internal priors or acquire new evidence, ensuring that the semantic gradients applied to the context are grounded in the external world.

\subsection{On the Pitfalls of the Sequential Training Pipeline}
Standard context training typically employs a linear, greedy strategy. 
It retains a single context $C_{t+1}$ at each training step and updates it based on a given batch. Nevertheless, as evidenced by our preliminary study on low-resource machine translation (specifically, translating English into Chokwe and Buginese), simply incorporating these web-searching tools has new risks, as detailed below.

\paragraph{Context Pollution}

\begin{figure}[h]
\vskip -2pt
    \centering
    \includegraphics[width=0.75\linewidth]{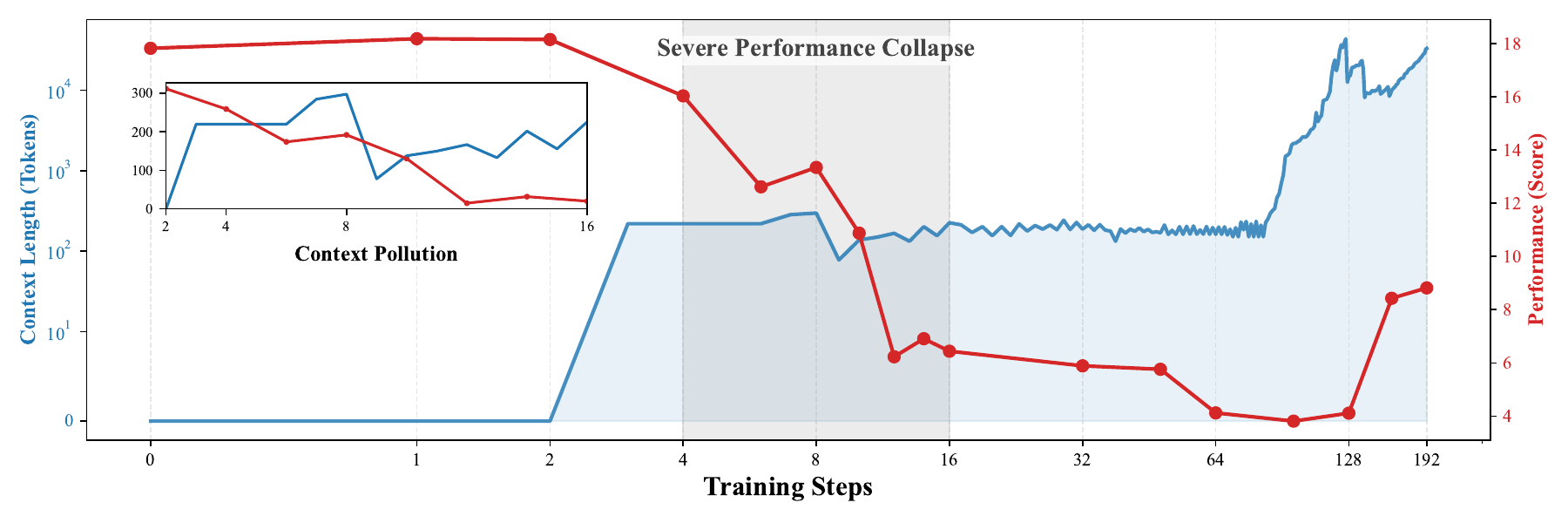}
    \caption{\footnotesize Context Pollution happens in the standard sequential context training: a tiny update could severely degrade context utility (steps 4 to 16), and the optimizer agent struggles to recover.}
    \label{fig:ctx_pollution}
\vskip -5pt
\end{figure}

Fig.~\ref{fig:ctx_pollution} illustrates our preliminary study on the \emph{English-to-Chokwe} translation task.
We observe that the context cocan be poisoned by tiny updates, resulting in a severe performance drop, and the optimizer agent struggles to remove these harmful artifacts once introduced.
As shown in the shaded region (steps 4 to 16), a mild update to the context (about 200 tokens) is associated with a precipitous decline in the performance score.
Crucially, the system fails to recover from this ``pollution'': Instead of pruning the toxic content, the optimizer repeatedly adds and removes information (steps 16 to 128) while the performance remains very low, highlighting the necessity of an explicit backtracking mechanism that helps the model to ``undo'' these kinds of mistakes.

\begin{figure}[htbp!]
\vskip -2pt
    \centering
    \includegraphics[width=0.8\linewidth]{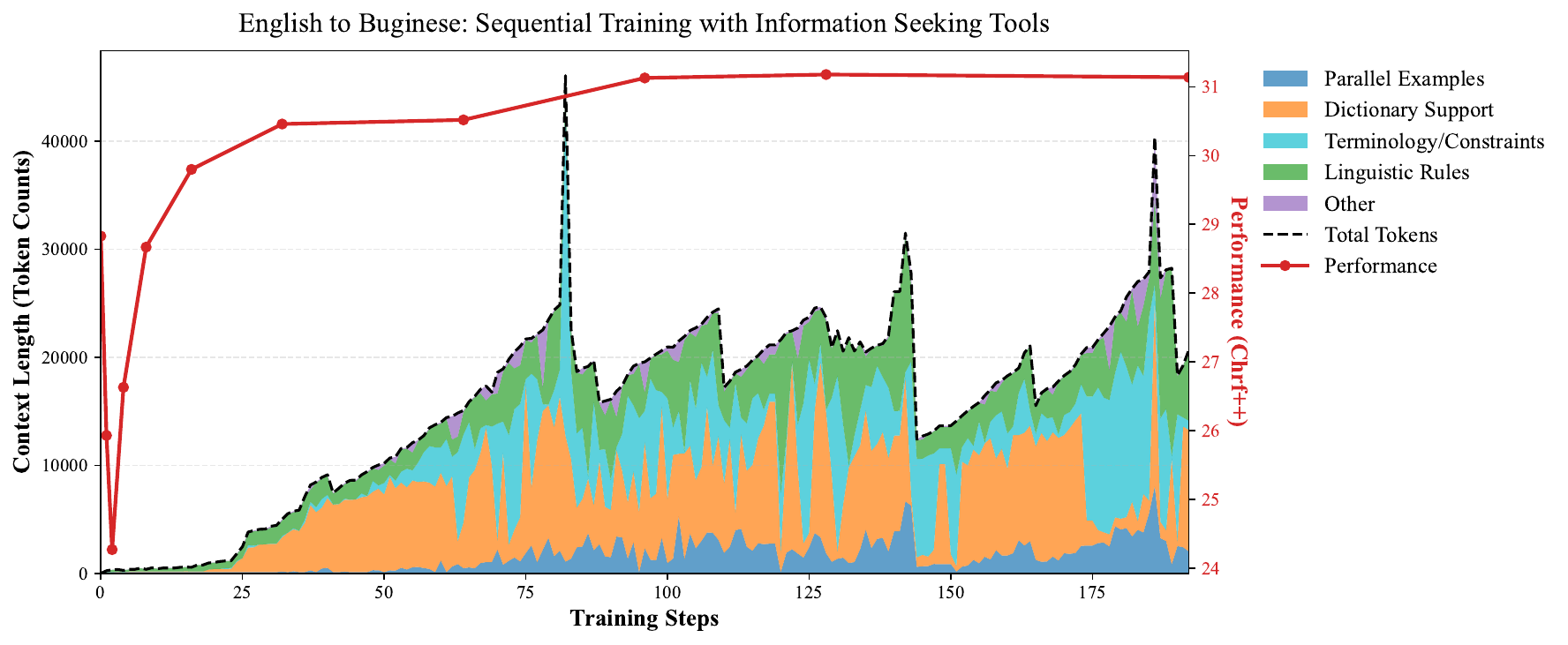}
    \caption{\footnotesize Standard sequential, linear training pipeline could suffer from being stuck at the local optima. As shown in the figure, the optimizer agent repeatedly adds and removes dictionary support.}
    \label{fig:local_optima}
    \vskip -5pt
\end{figure}

\paragraph{Local Optima}
The second flaw is that the model is prone to getting trapped in local optima, resulting in a repetitive cycle of accumulation and collapse.
As shown in Fig.~\ref{fig:local_optima} (English-to-Buginese translation task), the context length (dashed black line) exhibits a distinct sawtooth shape: it grows steadily before suffering sudden, sharp declines.
This behavior is reminiscent of the ``context collapse'' phenomenon observed in prior studies~\citep{DBLP:journals/corr/abs-2510-04618}, where models fail to maintain information density as length increases.
A closer inspection of the context composition reveals a more specific failure mode.
While the \textit{Dictionary Support} (orange region) consistently dominates the context, the optimizer does periodically attempt to prune these resources.
Yet, crucially, these pruned resources are invariably re-added in subsequent steps.
This implies that the optimizer is stuck in a loop: it tries to compress the context but fails to discover superior strategies (such as increasing \textit{Parallel Examples}, the blue region), and thus is forced to revert to the ``safe'' but suboptimal strategy of dictionary expansion.
This cyclical inability to escape the current strategy basin underscores the critical lack of effective exploration mechanisms in standard sequential training, especially when the context-optimizer agent has access to varying-quality external information.

\begin{figure}[h]
    \centering
    \includegraphics[width=0.9\textwidth]{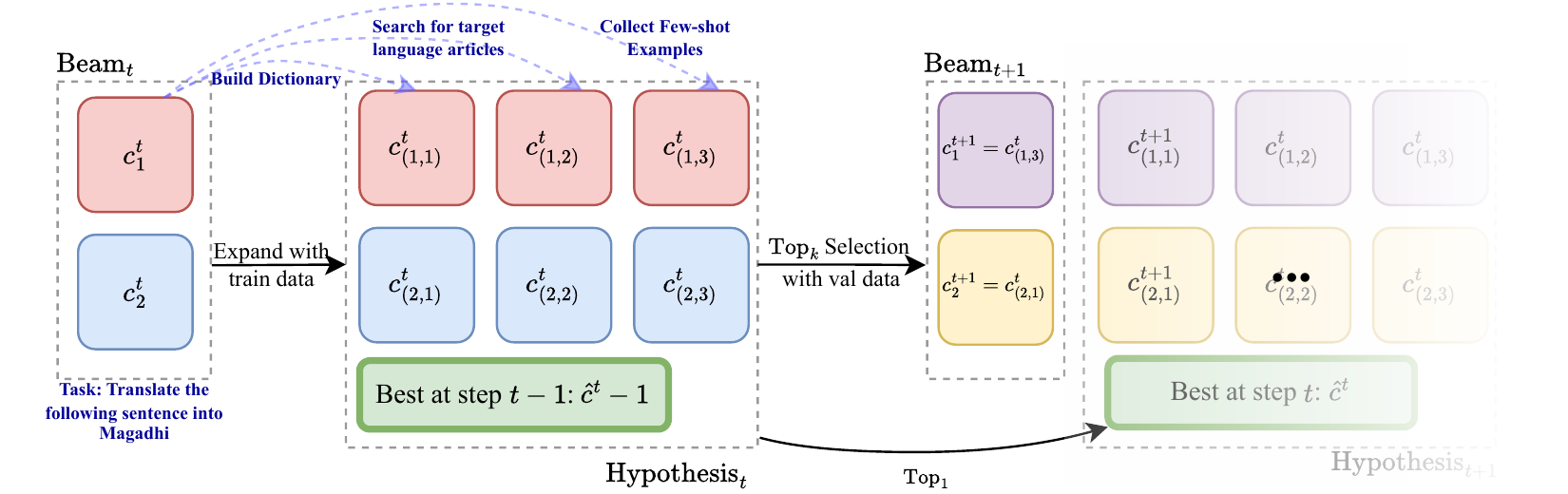}
    \caption{\footnotesize Beam Search-guided context training process. The optimizer agent could test different optimization strategies and reject lower-performing ones. We also include the best candidate from the previous step as the ``Do Nothing'' option.}
    \label{fig:context_training_bs}
\end{figure}
\subsection{Context Optimization Guided with Beam Search}
\label{sec:beam_search}

To address the two issues mentioned above, we adopt a beam-search-style training~\citep{vijayakumar2016diverse}. Instead of maintaining a single context trajectory and performing greedy sequential edits, we maintain a small population of $K$ candidate contexts, denoted by $\mathbb{C}_t$, and prune them using validation feedback:
\[
\mathbb{C}_t = \{c_t^{(1)}, c_t^{(2)}, \dots, c_t^{(K)}\}
\]
At each training step $t$, the pipeline proceeds in two phases (Fig.~\ref{fig:context_training_bs}):

\textbf{Expansion (Exploration):}
For each candidate in the current beam, the optimizer agent $\mathcal{O}$ generates $M$ child contexts by running the context-training loop in Fig.~\ref{fig:ctx_training} (dashed arrows in Fig.~\ref{fig:context_training_bs}). Each child is optimized for $L$ update steps on training batches sampled from $\mathcal{D}_{train}$, yielding multiple alternative context updates from the same parent. In machine translation, for example, different branches may emphasize building a dictionary, searching for reference articles, or collecting few-shot examples. These strategies are not hard-coded as a fixed menu; rather, they are discovered by the optimizer during expansion. To encourage diversity, children from the same parents are generated sequentially: When generating the context $c^t_{(i,j)}$, we provide the model with a short summary of the previous explorations $c^t_{(i,<j)}$ and explicitly prompt it to pursue a different update strategy.

\textbf{Selection (Pruning):}
We then evaluate all generated candidates, together with the best candidate from the previous step ($\hat{c}_{t-1}$, depicted as the green block), on a held-out validation set. The update rule is:
\begin{equation}
    \mathbb{C}_{t+1} = \text{Top}_K \left( \underbrace{\{\hat{c}_{t-1}\}}_{\text{Elitism}} \cup \bigcup_{c \in \mathbb{C}_t} \mathcal{O}_{\text{expand}}(c, \mathcal{B}_t) \right)
\end{equation}
where $\mathcal{O}_{\text{expand}}$ denotes the set of candidate contexts produced from $c$ through the expansion phase. This validation-guided pruning filters out branches that introduce noisy or harmful information via external tools before they can pollute the retained context. It also allows the search to abandon strategies that yield short-term progress but result in weaker validation performance. Including the best candidate from the previous step acts as a ``Do Nothing'' option: If all new explorations are unhelpful, the optimizer simply keeps the previous best state.

\textbf{Implementation via Version Control:}
To operationalize this branching pipeline, we implement the context as a version-controlled code repository. To that end, the optimizer uses atomic functions to manage context versions, such as \texttt{create\_branch} (to fork the current context for further expansion), \texttt{commit} (to snapshot a specific state of the context), and \texttt{check\_out} (to revert to a parent node or switch branches). These version-control actions are an implementation mechanism rather than a conceptual component of the method, and are currently hard-coded into the training loop. A more detailed description is shown in the Tab.~\ref{tab:tool_actions}.

\section{Experiments}

\subsection{Experiment Settings}
\paragraph{Datasets and Models}
To rigorously evaluate the proposed framework, we curate a diverse set of tasks spanning multiple domains and difficulty levels.
Our selection criteria are twofold: (1) ensuring broad coverage of distinct capabilities to demonstrate the generalizability of our approach; and (2) targeting tasks that \textit{a priori} benefit from external knowledge augmentation --- whether through linguistic resources, technical documentation, or domain-specific databases.
Specifically, we conduct experiments on the following benchmarks:

\begin{itemize}
    \item \textbf{Translating English to Low-resource Language (FLORES+~\citep{nllb-24,goyal-etal-2022-flores}):}
    % We select five low-resource languages from the FLORES++ dataset~\citep{nllb-24,goyal-etal-2022-flores}.
    This setting tests the model's ability to retrieve and use fundamental linguistic resources, such as dictionaries, grammar books, and few-shot parallel examples, to bridge the knowledge gap in the target language~\citep{DBLP:conf/iclr/TanzerSVJM24}. We select five languages where our base model \texttt{Gemini-2.5-Flash} does not perform well and that are not directly supported by Google Translate: Buginese, Magahi, Kikuyu, Chokwe, and Southwestern Dinka.

    \item \textbf{Clinical Scenario (HealthBench):}
    We use HealthBench~\citep{DBLP:journals/corr/abs-2505-08775} to evaluate the model's ability to perform realistic healthcare interactions.
    HealthBench simulates multi-turn conversations grounded in comprehensive, physician-written rubrics.
    The core challenge lies not only in possessing authoritative medical knowledge, but also in interacting in a manner consistent with physician experts.
    This setting effectively tests whether our method enables the agent to retrieve medical domain knowledge, verify clinical protocols, and dynamically adjust the executor model's behavior to meet expert standards.

    \item \textbf{Complex Reasoning \& Competitive Coding (HLE \& LiveCodeBench):}
    We include reasoning-heavy tasks like LiveCodeBench~\citep{DBLP:conf/iclr/JainHGLYZWSSS25} and Humanity's Last Exam (HLE)~\citep{phan2025humanitysexam}.
    These benchmarks demand that the agent actively search to bridge gaps in the model's parametric knowledge. While FLORES+ and HealthBench primarily assess the model's ability to retrieve and incorporate domain-specific information, we expect the LLM to be more familiar with these reasoning tasks because they are closer to its post-training distribution. Nevertheless, we hypothesize that seeking and incorporating external information may still be beneficial in these settings.
\end{itemize}
To simulate realistic deployment scenarios where a large amount of labeled data is often unavailable, we adopt a strictly constrained data setting.
For FLORES+, HealthBench, and LiveCodeBench, we limit the optimization budget to only \textbf{128 training samples} and \textbf{64 validation samples}. 
For HLE, different subdomains adopt different numbers of examples (around one hundred). The split details are in the App.~\ref{app:exp_details}. 
This low-resource regime serves as a rigorous measure of learning efficiency, challenging the agent to generalize well to new tasks with minimal supervision.
Unless otherwise specified, we employ \texttt{Gemini-2.5-Flash} as the backbone model for all experiments.
All other details about the data and the training are summarised in App.~\ref{app:exp_details}.

\paragraph{Evaluation Metrics}
For FLORES+, we report \textit{ChrF++} scores on the full test split. For HealthBench, we use the official rubric-based score. For LiveCodeBench, we report pass@1 and pass@8, and for HLE, we report average@8 accuracy. These metrics are computed on held-out test sets that are not used during context optimization.

\paragraph{Compared methods}
We validate our framework by comparing the performance of the following methods:
(1) Base LLM: The standard zero-shot performance of the backbone model without context training.
(2) Best-of-N (BoN): 
%A strong heuristic baseline that samples $N$ independent responses for data points in the training and validation set, and selects the best ones to compose the context. This serves as a non-iterative, brute-force reference.
This context comprises the model's best-of-$n$ responses ($n=8$) from the training and validation sets. This serves as a heuristic, non-iterative baseline.
(3) Sequential Training (Seq): The standard context training approach adopted by most prior works (e.g., OPRO), where the context is updated linearly based on the feedback on the training data point. We use the validation set to select the best ``checkpoint''.
(4) BeamSearch: Our proposed method that maintains a population of contexts to enable exploration.
Furthermore, we extend the Seq and BeamSearch methods with information-seeking capabilities (denoted as Seq-IS and BeamSearch-IS), where the optimizer is equipped with external tools.
To ensure a fair comparison, we align the computational budget across all optimization methods (BoN, Seq, and BeamSearch) by keeping the total number of calls to the optimizer agent roughly constant across different methods.

\subsection{Main results}

\paragraph{Results on Low-resource Translation}
Table~\ref{tab:results_mt} presents a performance comparison on the Low-Resource Machine Translation (LRMT) task, spanning English to five diverse low-resource languages. The empirical results yield several key insights:
First, our findings provide further evidence of the \textbf{Context Pollution} phenomenon. As shown in the table, directly equipping the standard sequential model with information-seeking tools (Seq-IS) consistently degrades performance across most language pairs compared to its closed variant (Seq).
Specifically, the average score drops from 31.13 (Seq) to 29.68 (Seq-IS). This confirms that, without a robust verification or backtracking mechanism, the introduction of external web-based information can introduce detrimental noise that eclipses the potential benefits
\begin{wraptable}{r}{0.625\textwidth}
\vspace{-10pt}
    \centering
    \small
    \caption{\footnotesize Results on low-resource machine translation demonstrate that, with the trained context, Gemini-2.5-Flash outperforms Gemini-2.5-Pro.}
    \label{tab:results_mt}
    \renewcommand{\arraystretch}{1.2}
    \setlength{\tabcolsep}{1.5pt}
    
    \resizebox{\linewidth}{!}{
        \begin{tabular}{lccccc>{\columncolor{yellow!20}}c}
        \toprule
        \multirow{2}{*}{\textbf{Methods}} 
        & \multicolumn{6}{c}{\textbf{Low Resource Machine Translation}} \\
        \cmidrule(lr){2-7}
         & \textbf{bug\_latn} & \textbf{mag\_Deva} & \textbf{kik\_Latn} & \textbf{cjk\_latn} & \textbf{dik\_Latn} & \textbf{Avg.} \\
        \midrule
        Gemini-2.5-Flash & 28.83 & 44.86 & 34.43& 17.83 & 5.62 & 26.31\\
        Gemini-2.5-Pro & 31.42 & 42.42 & 35.89 & 22.89 & 19.21 & 30.37\\
        \midrule
        \multicolumn{7}{c}{Context Training} \\
        \midrule
        BoN & 32.84 & 47.83 & 37.17 & 23.72 & 18.12 & 31.94\\
        Seq    & 32.53 & 46.01& 36.34& 24.49& 18.25 & 31.13\\
        BeamSearch & 32.68 & 46.55& 36.46& 22.47 & 18.41 & 31.31 \\
        \midrule
        \multicolumn{7}{c}{Context Training with Information Seeking} \\
        \midrule
        Seq-IS  & 31.15  & 45.58& 37.53& 18.16 & 15.96 & 29.68\\
        BeamSearch-IS & \textbf{33.74} & \textbf{50.52} & \textbf{39.73} & \textbf{26.25} & \textbf{22.46}&  \textbf{34.51}\\
        \bottomrule
        \end{tabular}
    }
\vspace{-10pt}
\end{wraptable}
of the retrieved context.
In contrast, our proposed BeamSearch-IS
 effectively mitigates this issue, achieving a substantial performance leap to an average of \textbf{34.51}. This represents an improvement of \textbf{4.14 points} over the best non-IS context training baseline (BoN), demonstrating that our method successfully discriminates between high-quality evidence and noisy retrievals. Notably, BeamSearch-IS not only surpasses its base models but also outperforms the much larger \textbf{Gemini-2.5-Pro} baseline (30.37) in our evaluation.

\begin{figure}[t]
    \centering
    \includegraphics[width=0.9\textwidth]{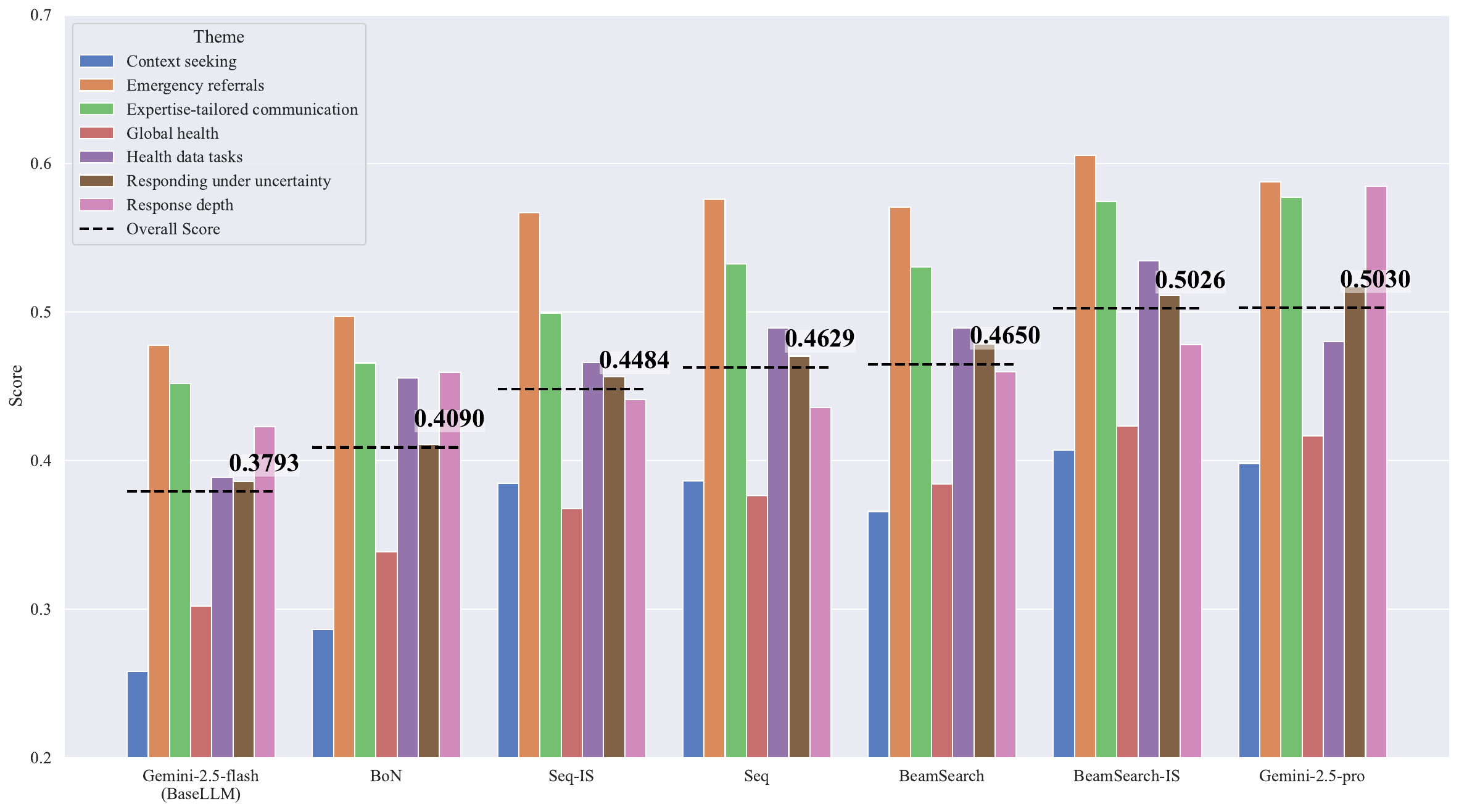}
    \caption{\footnotesize Performance comparison on the HealthBench dataset across different models and methods. The overall score (dashed line) and theme-specific scores demonstrate the efficacy of our BeamSearch-IS approach compared to baselines.
    }
    \label{fig:healthbench_results}
\vskip -15pt
\end{figure}

\paragraph{Results on HealthBench}
We present the comparative results on HealthBench in Fig.~\ref{fig:healthbench_results}. The trends observed here reinforce our findings from the translation task.
First, the phenomenon of Context Pollution is evident: simply augmenting the sequential model with external tools (Seq-IS) proves detrimental, yielding an overall score of 0.4484, which underperforms the tool-free Seq baseline (0.4629).
Our proposed BeamSearch-IS effectively mitigates this noise, achieving an overall score of \textbf{0.5026}. Notably, BeamSearch-IS achieves performance comparable to the much larger and more expensive \texttt{Gemini-2.5-Pro} (0.5030).
When diving deeper into performance on different themes, our method performs well in \textbf{Health Data Tasks} (handling health data accurately) and even outperforms the Pro model in \textbf{Emergency Referrals} (recognizing emergencies
and steering people toward care), suggesting that, in this setting, active context verification can be more effective than simply scaling the model for rigid, error-sensitive requirements.
In contrast, \texttt{Gemini-2.5-Pro} retains a clear lead in \textbf{Response Depth} (whether
models can adjust the depth of their responses to match user needs), indicating that, while our pipeline improves accuracy and recognition of emergencies, the intrinsic generation capability of larger models remains a distinct advantage.

\begin{table*}[h]
    \small
    \centering
    \caption{\footnotesize \textbf{Results on Complex Reasoning Tasks.} We report pass@1 / pass@8 (\%) for LiveCodeBench and average@8 accuracy (\%) for HLE. Our BeamSearch-IS achieves the best balance across both domains, outperforming baselines in both coding and multidisciplinary reasoning.}
    \label{tab:complex_reasoning_merged}
    \renewcommand{\arraystretch}{1.2}
    \setlength{\tabcolsep}{3.5pt}
    
    \resizebox{0.85\textwidth}{!}{
        \begin{tabular}{l | cc >{\columncolor{yellow!20}}c | ccccc >{\columncolor{yellow!20}}c}
        \toprule
        \multirow{2}{*}{\textbf{Methods}} 
        & \multicolumn{3}{c|}{\textbf{LiveCodeBench (pass@1 / pass@8)}} 
        & \multicolumn{6}{c}{\textbf{HLE Accuracy (\%)}} \\
        \cmidrule(lr){2-4} \cmidrule(lr){5-10}
         & \textbf{Medium} & \textbf{Hard} & \textbf{Overall} 
         & \textbf{Bio} & \textbf{CS} & \textbf{Phys} & \textbf{Math} & \textbf{Hum} & \textbf{Avg.} \\
        \midrule
        Gemini-2.5-Flash & 71.5 / 83.8 & 30.0 / 49.6 & 49.4 / 65.6 & 7.10 & 6.46 & 5.00 & 8.08 & 6.01 & 6.53 \\
        \midrule
        \multicolumn{10}{c}{\textit{Context Training}} \\
        \midrule
        BoN              & 72.5 / 83.6 & 28.7 / 50.3 & 49.2 / 65.9 & 6.25 & 4.21 & 4.06 & 10.13 & 6.01 & 6.13 \\
        Seq              & 68.9 / 85.6 & 31.5 / 51.6 & 49.0 / 67.5 & 7.24 & 2.81 & 6.88 & 8.08 & 6.01 & 6.20 \\
        BeamSearch       & 69.9 / 84.2 & 31.2 / 52.0 & 49.3 / 67.1 & 7.39 & 3.90 & 5.94 & 8.97 & 6.33 & 6.51 \\
        \midrule
        \multicolumn{10}{c}{\textit{Context Training with Information Seeking}} \\
        \midrule
        Seq-IS           & 71.6 / 86.2 & 29.6 / 52.1 & 49.3 / 68.1 & \textbf{8.81} & 6.74 & 3.22 & 6.54 & 6.96 & 5.38 \\
        BeamSearch-IS    & \textbf{73.5 / 89.0} & \textbf{33.9 / 57.2} & \textbf{52.5 / 70.2} & \textbf{8.81} & \textbf{8.30} & \textbf{7.67} & \textbf{11.15} & \textbf{7.23} & \textbf{8.63} \\
        \bottomrule
        \end{tabular}
    }
    %\vskip -10pt
\end{table*}

\paragraph{Results on Complex Reasoning Tasks}
Table~\ref{tab:complex_reasoning_merged} summarizes the performance on LiveCodeBench (LCB) and the Humanity's Last Exam (HLE). We observe three key trends:
(1) Methods relying solely on internal reasoning (BoN, Seq, BeamSearch) yield negligible improvements over the \texttt{Gemini-2.5-Flash} baseline. The overall pass@1 on LCB hovers around 49\%, and HLE accuracy remains approximately 6.5\%, indicating that optimizing reasoning paths without external information yields limited gains in our setting.
(2) The Seq-IS method shows inconsistent results across domains. While it remains robust on LCB (improving pass@8 to 68.1\%), its performance degrades on HLE (dropping to 5.38\%), particularly in reasoning-heavy subjects like Physics and Math.
(3) In contrast, BeamSearch-IS achieves consistent gains across both benchmarks. It improves pass@1 on LCB Hard to 33.9\% (vs 30.0\% baseline) and achieves the highest average accuracy on HLE (8.63\%), effectively outperforming both the baseline and other context training methods.

\section{Analysis}

\begin{figure}[t]
    \centering
    \includegraphics[width=\textwidth]{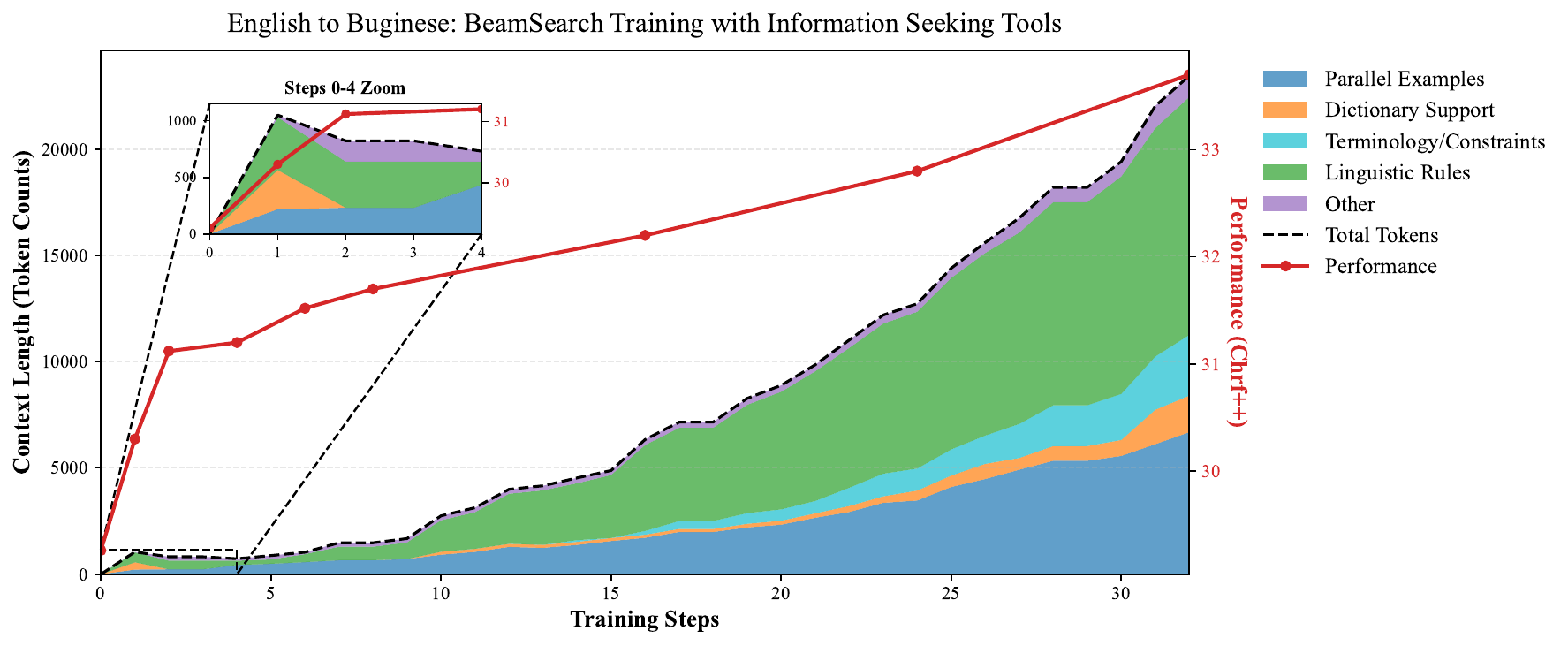}
    \vskip -10pt
    \caption{\footnotesize We visualize the composition of contexts with different resource types along the training and the corresponding translation performance (red line) on the English-to-Buginese task. The zoom-in window (top-left) highlights a critical "self-correction" mechanism: the model initially explores \textit{Dictionary Support} (orange) at Step 1 but quickly discards it in favor of \textit{Linguistic Rules} (green) and \textit{Parallel Examples} (blue), effectively avoiding the local optima.}
    \vskip -5pt
    \label{fig:local_optima_bs}
\end{figure}
\paragraph{How Beam Search Enables Exploration}
Fig.~\ref{fig:local_optima_bs} illustrates how Beam Search helps navigate the optimization landscape and avoid local optima. Compared with the sequential baseline in Fig.~\ref{fig:local_optima}, which tends to remain on less effective strategies such as \textit{Dictionary Support}, the Beam Search trajectory moves toward a different combination of resources.
As shown by the expanding green and blue regions in the main plot, the optimizer identifies that combining \textit{Linguistic Rules} with \textit{Parallel Examples} yields better translation quality than a simple vocabulary expansion, consistent with recent observations from~\citet{DBLP:conf/iclr/AycockSWMS25}.
This behavior is most visible in the zoom-in window (Steps 0--4). At Step 1, the model tentatively increases its use of \textit{Dictionary Support} (the orange spike). Rather than committing to this branch, Beam Search evaluates it against alternatives and discards it in Step 2, shifting toward a combination of linguistic rules and parallel examples. This ability to test and reject candidate strategies supports broader exploration and avoids the early saturation seen in the sequential baseline.
\paragraph{Data Efficiency Analysis}
\begin{figure}[t]
    \centering
    \captionsetup[subfigure]{justification=centering,singlelinecheck=false}
    \begin{subfigure}[t]{0.48\linewidth}
        \centering
        \includegraphics[width=\linewidth]{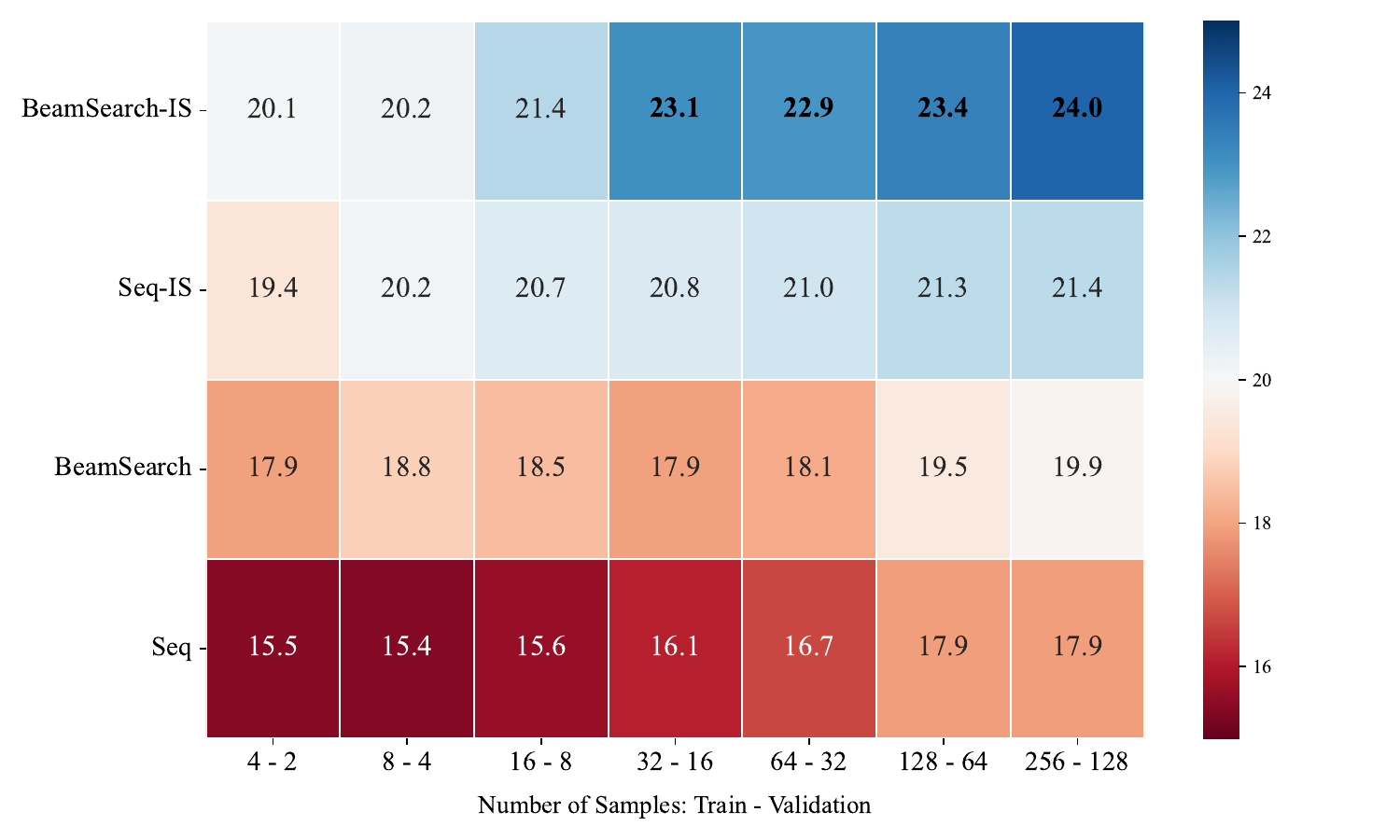}
        \caption{\footnotesize Data efficiency.}
        \label{fig:data_efficiency}
    \end{subfigure}
    \hfill
    \begin{subfigure}[t]{0.48\linewidth}
        \centering
        \includegraphics[width=\linewidth]{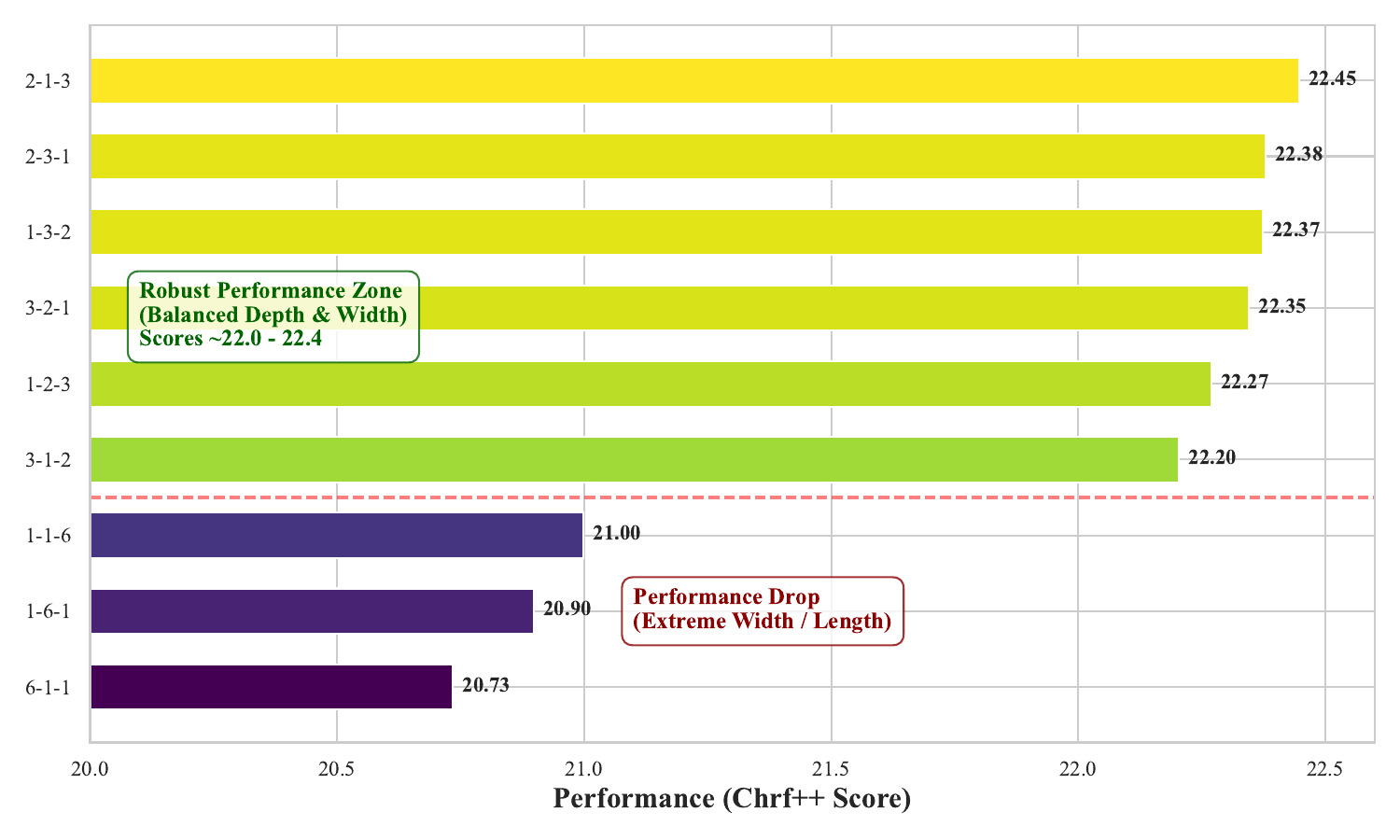}
        \caption{\footnotesize Hyperparameter ablation.}
        \label{fig:beam_search_hp}
    \end{subfigure}
    \caption{\footnotesize \textbf{Data efficiency and hyperparameter robustness.} (a) The heatmap compares different methods across varying training data sizes (from 4 to 256 samples) on the English-to-Southwestern Dinka (dik\_Latn) translation task. Darker blue indicates higher scores, while red hues denote lower performance. BeamSearch-IS rapidly converges to high-performance regions (score $>23.0$) with as few as 32 samples, highlighting its superior sample efficiency. (b) We report Chrf++ scores across different training configurations. The system exhibits a wide "Robust Performance Zone" (yellow).}
    \label{fig:analysis_heatmaps}
    \vskip -15pt
\end{figure}
Acquiring large-scale annotated data is often prohibitively expensive in real-world scenarios. Therefore, we analyze performance across varying data regimes, ranging from $4$ to $256$ training samples, to evaluate the practicality of our approach.
We chose the English-to-Southwestern Dinka (dik\_Latn) translation task for this analysis.
Figure~\ref{fig:data_efficiency} illustrates a clear performance hierarchy. While standard methods (e.g., Seq, bottom rows) remain stuck in low-performance regions (indicated by red hues) even as data volume increases, our BeamSearch-IS method demonstrates remarkable data efficiency. Notably, it achieves near-optimal performance (score $>23.0$, dark blue) with as few as 32 training samples. This suggests that Beam Search acts as a \textit{signal amplifier}: by exploring multiple potential context modifications for each training example, the optimizer extracts more signal from limited data, allowing it to converge to high-quality strategies with a fraction of the data required by baseline methods.

\paragraph{Ablation Study on Hyper-parameters}
We further assess the framework's sensitivity to the hyper-parameters of Beam Search: Beam Width, Number of Hypotheses per step, and training epochs. We again use the English-to-Southwestern Dinka (dik\_Latn) translation task for this ablation study. Figure~\ref{fig:beam_search_hp} illustrates the performance variance across distinct combinations under a similar training budget, denoted as ({Width-Hypotheses-Epochs}). As highlighted by the "Robust Performance 
Zone" (yellow bars), the system maintains high performance ($\sim$22.2--22.45) across a diverse range of balanced configurations (e.g., 2-1-3 or 3-2-1). 
This indicates that BeamSearch-IS is relatively insensitive to specific parameter settings. Significant performance degradation occurs only in "extreme, unbalanced settings" (purple bars), such as the 6-1-1 configuration (prioritizing extreme width over training epochs), where the score drops to 20.73. This confirms that as long as a reasonable balance between exploration (width) and exploitation (depth) is maintained, the method yields stable results.

\paragraph{Data Utility Analysis}
\begin{wrapfigure}{R}{0.6\textwidth}
\vskip -10pt
    \centering
    \begin{subfigure}{0.48\linewidth}
        \centering
        \includegraphics[width=\linewidth]{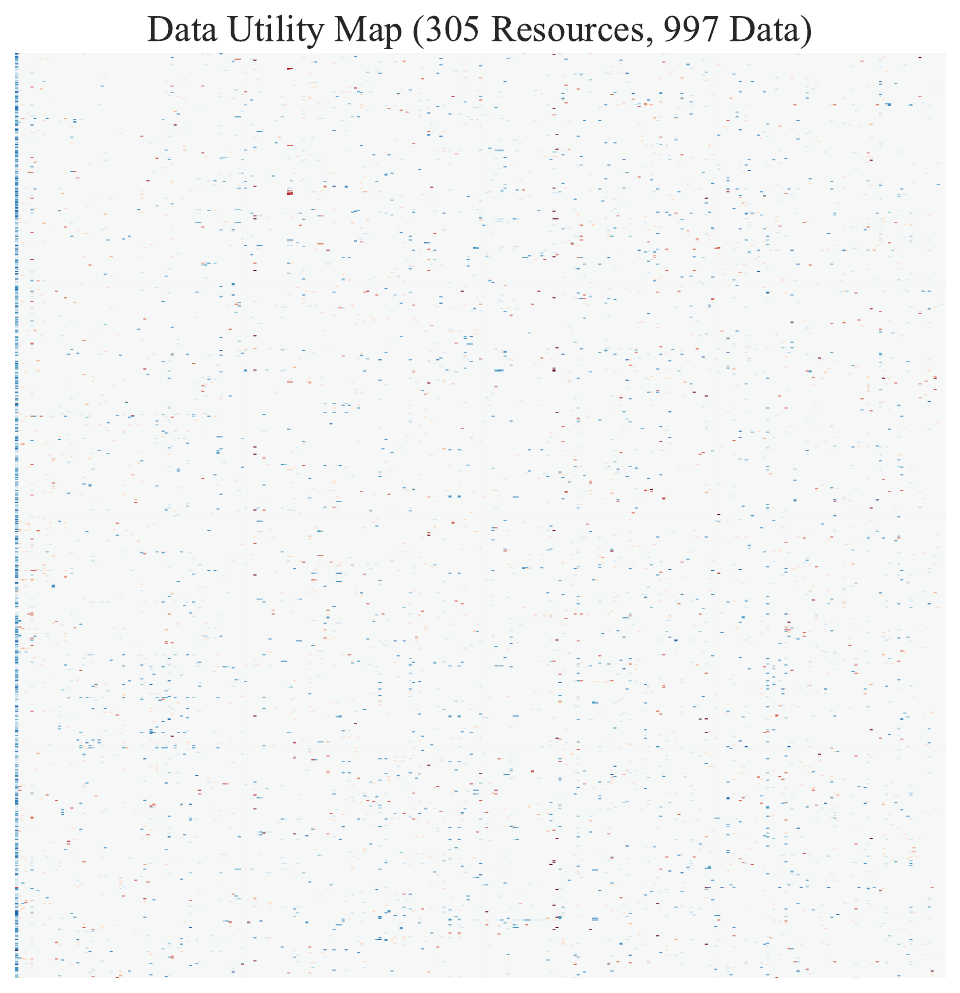}
        \caption{\footnotesize English to Magahi Translation}
    \end{subfigure}
    \hfill
    \begin{subfigure}{0.48\linewidth}
        \centering
        \includegraphics[width=\linewidth]{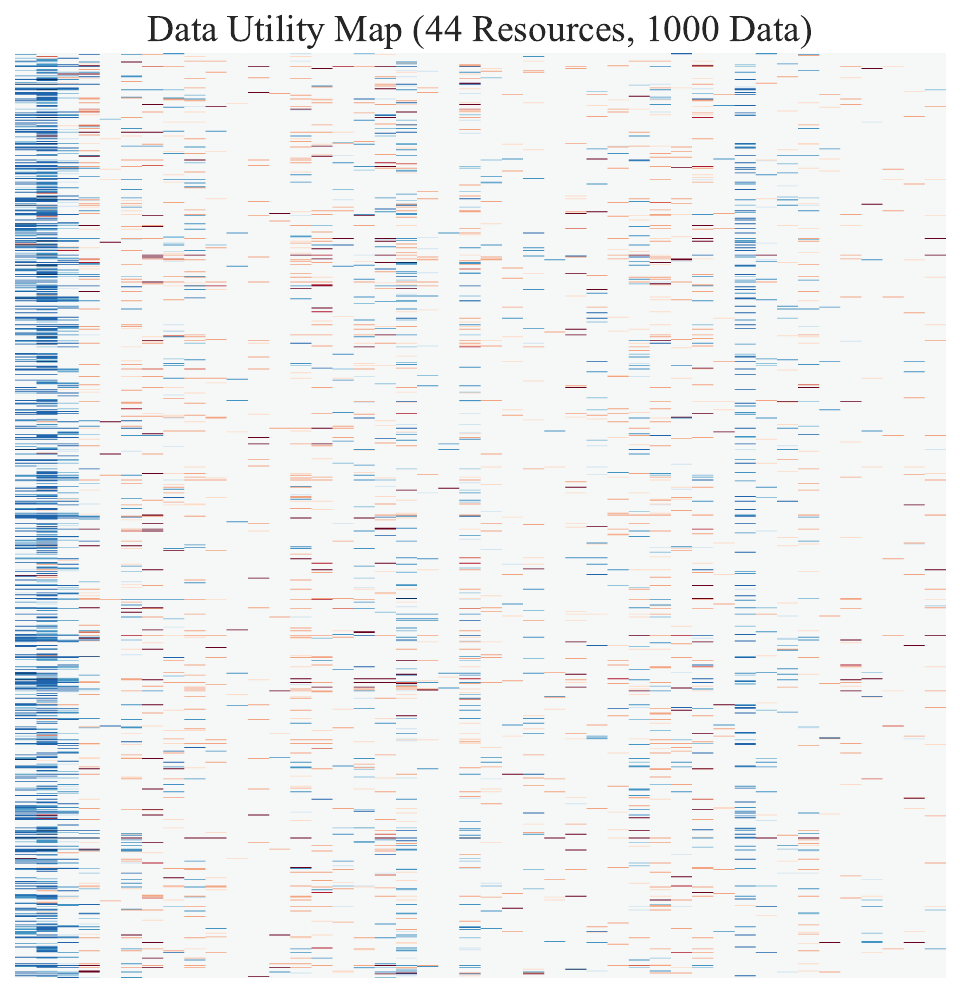}
        \caption{\footnotesize HealthBench}
    \end{subfigure}
    \caption{\footnotesize The heatmaps visualize the utility of retrieved resources (x-axis) for test samples (y-axis). \textcolor{blue}{Blue} indicates positive transfer. While the overall utility map is sparse, distinct vertical striations appear in both tasks, revealing the existence of "dominant resources" that provide universal benefits across diverse data points.}
    \label{fig:utility_analysis}
    \vskip -5pt
\end{wrapfigure}
To better understand how the context works, we use \texttt{Gemini-3-Flash} as an annotator to quantify the utility of each resource for each test sample. The specific prompt we used for annotation is in App.~\ref{app:prompts}. Figure~\ref{fig:utility_analysis} visualizes this utility map, where the x-axis represents unique context resources (sorted from longer to shorter ones from left to right) and the y-axis represents test data points.
Our observations include:
(1) The overall maps are predominantly sparse and modular, indicating that the majority of retrieved contexts are highly instance-specific and only effective for a narrow subset of queries.
(2) We also observe that dominant resources emerge in both domains, visualized as distinct continuous vertical blue lines on the left. These lines correspond to a small set of high-quality contexts that provide positive transfer across nearly the entire test set.
(3) Crucially, to ensure these dominant resources were not merely leaking answers, we rigorously screened them for data contamination (defined as the presence of a question and its ground-truth answer). We again use \texttt{Gemini-3-Flash} as the data contamination detector; the specific prompt is in App.~\ref{app:prompts}. Our inspection confirmed \textbf{zero instances of such overlap}, ensuring that the broad utility of these resources stems from genuine knowledge applicability rather than data leakage.

\begin{wraptable}{h}{0.75\textwidth}
    \centering
    \small
    \caption{\footnotesize Applying context optimized by Gemini-2.5-Flash to the stronger Gemini-3-Flash. BeamSearch-IS improves the performance of the stronger model across all domains.}
    \label{tab:transferability}
    \renewcommand{\arraystretch}{1.2}
    \setlength{\tabcolsep}{2pt}
    
    \resizebox{\linewidth}{!}{
        \begin{tabular}{l cc ccccc c}
        \toprule
        \multirow{2}{*}{\textbf{Methods}} 
        & \multicolumn{2}{c}{\textbf{Low-Resource MT}} 
        & \multicolumn{5}{c}{\textbf{HLE Subjects (Acc \%)}} 
        & \multirow{2}{*}{\textbf{HealthBench}} \\
        \cmidrule(lr){2-3} \cmidrule(lr){4-8} 
         & \textbf{Buginese} & \textbf{Magahi} 
         & \textbf{Bio/Med} & \textbf{CS/AI} & \textbf{Physics} & \textbf{Math} & \textbf{Humanity} 
         & \\
        \midrule
        Gemini-3-Flash & 32.35 & 42.80 & 26.70 & 27.78 & 29.06 & 43.08 & 39.56 & 0.6164 \\
        \midrule
        Seq            & 32.61 & 46.46 & 26.98 & 25.56 & 29.06 & 41.28 & 41.14 & 0.6011 \\
        BeamSearch     & 33.50 & 46.80 & 26.99 & 27.08 & 25.00 & 43.33 & 43.67 & 0.6218 \\
        Seq-IS         & 33.28 & 46.57 & 27.86 & 28.08 & 27.81 & 44.36 & 42.41 & 0.6118 \\
        \midrule
        \rowcolor{yellow!15} 
        \textbf{BeamSearch-IS} & \textbf{34.40} & \textbf{52.12} & \textbf{31.25} & \textbf{32.58} & \textbf{29.38} & \textbf{44.99} & \textbf{46.20} & \textbf{0.6624} \\
        \bottomrule
        \end{tabular}
    }
\vskip -15pt
\end{wraptable}
\paragraph{Generalization of the trained context}
Finally, we investigate a critical question: Does the trained context discover genuinely helpful knowledge, or does it merely learn model-specific artifacts? To test this, we evaluate the \textit{transferability} of the context by applying context optimized on \texttt{Gemini-2.5-Flash} directly to the newer, more capable \texttt{Gemini-3-Flash} without \emph{any} further tuning.
Table~\ref{tab:transferability} reveals that the standard closed Seq method exhibits poor transferability and yields no performance gain in most cases.
For instance, Seq causes regression on HealthBench (a slight drop from 0.6164 to 0.6011) and struggles with reasoning-heavy HLE subjects like CS/AI and Math. This suggests that these baselines tend to learn executor-specific patterns that do not generalize.
Conversely, BeamSearch-IS demonstrates remarkable generalization. It drives consistent gains across all tasks, boosting the Magahi translation score \textbf{by nearly 10 points} (to \textbf{52.12}) and achieving \textbf{0.6624} on HealthBench. Notably, in the HLE benchmark, it consistently delivers gains across diverse fields, and those gains are in fact \emph{even larger} than on the \texttt{Gemini-2.5-Flash} model. This confirms that our context optimization approach can retrieve and leverage high-quality, generalizable, and model-agnostic knowledge that helps even stronger models perform better.

\section{Conclusion}

Recent work employing LLMs as context optimizers suggests a promising path for specializing models at test time without updating their weights.
However, most existing context optimization methods are passive and closed, which limits their ability to acquire missing external knowledge.
In this paper, we study context optimization with active information seeking by augmenting the optimizer with Wikipedia search and browser-based tools. We show that naively adding these tools to sequential context training can be harmful, but that a simple search-based training procedure makes them effective in practice.
Across low-resource machine translation, healthcare, and reasoning benchmarks, this combination yields consistent improvements over passive baselines, while also remaining data-efficient and transferring well across models.

\section{Limitations and Future Work}
The approach we investigate in this paper has the following two limitations:
(1) The method relies on the \textbf{context utilization} ability of the base model. Comparing HLE results in Tab.~\ref{tab:complex_reasoning_merged} and Tab.~\ref{tab:transferability}, we observe that although context is trained with \textit{Gemini-2.5-Flash}, it yields larger performance gains when applied to the more capable \textit{Gemini-3-Flash} model. This suggests that while our optimizer agent successfully retrieves useful resources from the web, the executor agent may be unable to use the resulting context effectively for problem-solving.
(2) Fig.~\ref{fig:utility_analysis} reveals the sparsity of the constructed context: Apart from a few resources that are generally helpful (like general guidelines), the majority of collected resources are pointwise and highly data-specific. This indicates that if a task is highly instance-specific, the training set may fail to capture the diversity of the test distribution, making the optimized context difficult to generalize. This observation also contextualizes the success of prior work on agentic tasks or games, where knowledge and action patterns are highly reusable, in contrast to the distinct, one-off reasoning required in our domain.

In response, future work could focus on: (1) enhancing the context utilization ability of base models to maximize the utility of retrieved contexts; and (2) equipping agents with broader exploration capabilities (i.e., wide search) to gather more comprehensive and diverse information and mitigate sparse or biased context retrieval. Another direction is to investigate hybrid offline \& online settings, where models combine offline preparation of general background knowledge with online exploration for instance-specific information.

\bibliography{main}

\clearpage
\section{Appendix}

\subsection{Algorithm}

\begin{algorithm}[h]
\caption{Beam-Search-Guided Context Training}
\begin{algorithmic}[1]
\State \textbf{Input:} Train Data $\mathcal{D}_{train}$, Validation Data $\mathcal{D}_{val}$, Beam Width $K$, Branching Factor $M$, Optimization Steps per Child $L$
\State \textbf{Initialize:} Empty Context $C_0$, $s_0 \leftarrow \text{Validate}(C_0, \mathcal{D}_{val})$, Beam set $\mathbb{C} \leftarrow \{ C_0 \}$, $(C_{\text{best}}, s_{\text{best}}) \leftarrow (C_0, s_0)$
\While{Global Step $<$ Max Steps}
    \State $\mathcal{C}_{candidates} \leftarrow \{ (C_{\text{best}}, s_{\text{best}}) \}$ \Comment{Include the current global best}

    \State \textbf{Phase 1: Expansion \& Optimization}
    \For{each parent context $C_k \in \mathbb{C}$}
        \For{$i \leftarrow 1$ to $M$}
            \State Copy $C_k$ as $C_k^i$
            \For{optimization step $l \leftarrow 1$ to $L$} \Comment{Do $L$ steps optimization}
                \State Sample task batch $X_l \sim \mathcal{D}_{train}$
                \State $\hat{Y}_l^i \leftarrow \text{Executor}(X_l, C_k^i)$
                \State $r_l^i \leftarrow \text{Reward}(X_l, \hat{Y}_l^i)$
                \State $\mathcal{B}_l^i \leftarrow (X_l, \hat{Y}_l^i, r_l^i)$ \Comment{Construct a learnable batch}
                \State $C_k^i \leftarrow \text{OptimizerAgent}(\mathcal{B}_l^i, C_k^i)$ \Comment{Update the context}
            \EndFor
            
            %\State \Comment{Evaluate Candidate}
            \State $\text{score}_k^i \leftarrow \text{Validate}(C_k^i, \mathcal{D}_{val})$
            \State $\mathcal{C}_{\text{candidates}} \leftarrow \mathcal{C}_{\text{candidates}} \cup \{ (C_k^i, \text{score}_k^i) \}$
        \EndFor
    \EndFor

    \State \textbf{Phase 2: Selection with Elitism}
    \State $\mathbb{C} \leftarrow \{ C \mid (C, s) \in \operatorname{Top}_K(\mathcal{C}_{candidates}) \}$
    \If{$\max_{(C, s) \in \mathcal{C}_{candidates}} s > s_{\text{best}}$}
        \State $(C_{\text{best}}, s_{\text{best}}) \leftarrow \operatorname{MaxScore}(\mathcal{C}_{candidates})$ \Comment{Update global best}
    \EndIf
\EndWhile
\end{algorithmic}
\end{algorithm}

\subsection{Experiment Details}
\label{app:exp_details}

This section provides more detailed experimental configurations, including data splits and training settings.

\paragraph{Data Split}
(1) FLORES++ provides the official \texttt{dev} (997 examples) and \texttt{devtest} split (1012 examples). For all five languages, we randomly sample 128 (64) samples for training (validation) from the \texttt{devtest} split, and use the entire \texttt{dev} split as the test set.
(2) The original HealthBench contains 5,000 problems. We randomly sample 128 problems for training, 64 problems for validation, and 1000 problems for testing.
(3) The LiveCodeBench (V6 release) contains 1055 problems from 5/1/2023 to 5/1/2025. We focus on medium and hard problems with timestamps from 5/1/2024 to 5/1/2025, and randomly sample 128 problems for training, 64 problems for validation, and 128 for testing.
(4) Humanity's Last Exam consists of 2,500 exam questions in over a hundred subjects, grouped here into 8 high-level categories: Math (41\%), Biology/Medicine (11\%), Computer Science/Artificial Intelligence (CS/AI, 10\%), Physics (9\%), Humanities/Social Science (*9\%), Chemistry (7\%), Other (9\%), and Engineering (4\%). During the preliminary study, we found that the knowledge required to solve these problems is highly instance-specific, even within the same category, resulting in very limited positive transfer between the training, validation, and test sets. We therefore use an LLM to score the data relevance between problems within the same category and make the training set representative enough. Finally, we investigate the following five domains. The data splits are detailed in the table:
\begin{table}[]
    \centering
    \begin{tabular}{c|ccc}
    \toprule
        Category & Train & Validation & Tests \\
    \midrule 
        Biology/Medicine & 90 & 44 & 88 \\
        CS/AI & 91 & 44 & 89 \\
        Physics & 82 & 40 & 80 \\
        Math & 148 & 78 & 780 \\
        Humanity & 95 & 19 & 79 \\
    \midrule
    \end{tabular}
    \caption{\footnotesize Data split details for HLE benchmark}
    \label{tab:placeholder}
\end{table}

\begin{table}[h]
\centering
\caption{Complete list of actions and descriptions for ContextManageTool.}
\label{tab:tool_actions}
\small
\renewcommand{\arraystretch}{1.25}
\begin{tabularx}{\textwidth}{@{} l |  l  | X @{}}
\toprule
\textbf{Category} & \textbf{Action} & \textbf{Description} \\
\midrule

% --- Resource Management ---
\multirow{9}{*}{\textbf{Context Edition}} 
 & \texttt{create} & Initializes a new empty context. \\
 & \texttt{add} & Injects a new atomic resource (text, URL, code) into the context. \\
 & \texttt{update} & Modifies specific fields of an existing resource. \\
 & \texttt{remove} & Deletes a specific resource from the current context. \\
 & \texttt{swap} & Exchanges the positions of two resources (for priority adjustment). \\
 & \texttt{merge} & Consolidates two different resources into one. \\
 & \texttt{set\_active} & Sets the focus to a specific context ID for subsequent operations. \\
 & \texttt{get\_resource} & Retrieves the full content and metadata of a specific resource. \\
 & \texttt{delete} & Permanently removes an atomic resource. \\
\midrule

% --- Retrieval & Search ---
\multirow{4}{*}{\textbf{Context Read}} 
 & \texttt{search} & Performs keyword-based filtering on resource content and tags. \\
 & \texttt{embedding\_search} & Retrieves resources using cosine similarity of vector embeddings. \\
 & \texttt{llm\_search} & Invokes a sub-agent to read, reason, and rank resources based on a complex query. \\
 & \texttt{list\_resources} & Lists resources in the active context with adjustable detail levels (Summary/Preview/Detail). \\

 \midrule

 % --- Version Control System ---
\multirow{7}{*}{\textbf{Version Control}} 
 & \texttt{create\_branch} & Forks the current context state into a new named branch. \\
 & \texttt{checkout} & Switches the working directory to a specific branch or commit. \\
 & \texttt{commit} & Creates an immutable snapshot of the current resource state. \\
 & \texttt{merge\_branch} & Merges commit history and resources from a source branch to a target. \\
 %& \texttt{diff} & Computes differences (added/removed/modified) between branches. \\
 & \texttt{list\_branches} & Displays all branches with metadata (head commit, description). \\
 & \texttt{update\_branch\_info} & Updates auxiliary metadata (e.g., validation scores) for a branch. \\

\bottomrule
\end{tabularx}
\end{table}

\paragraph{Training settings}
For BeamSearch-IS and BeamSearch, we set K (the beam width) to 2 and M (the branching factor) to 3 for all tasks. For machine translation, we train the context for 2 epochs; for other tasks, we train for 1 epoch.
To ensure fair comparison, we train Seq-IS and Seq for 12 epochs on FLORES and 6 epochs for other tasks.

\subsection{Context Management Tool}
\label{app:tool_actions}

Table~\ref{tab:tool_actions} provides the detailed description of all atomic actions in our implemented \texttt{ContextManageTool}, categorized by their different functionalities.

\subsection{Prompts}
\label{app:prompts}

% --- Box 1: Context Engineer System Prompt ---
\begin{promptbox}{Optimizer Agent System Prompt}
\textbf{\#\#\# ROLE \& MISSION}

You are an autonomous \textbf{Context Engineer} that optimizes a structured context for a given task for one step.

Your optimized context will finally be delivered to a downstream \textbf{Executor} Agent to perform the task. With this context, the Executor Agent should be more likely to succeed on similar tasks.

\vspace{0.5em}
\textbf{\#\#\# TOOL AUGMENTATION}

You are augmented with the following tools to manage the context:

\texttt{\{\%- for tool in tools.values() \%\}} \\
* \code{\{\{ tool.name \}\}}: \code{\{\{ tool.description \}\}} \\
\hspace*{1em} Takes inputs: \code{\{\{tool.parameters.properties\}\}} \\
\hspace*{1em} Returns an output of type: \code{\{\{tool.output\_type\}\}} \\
\texttt{\{\%- endfor \%\}}

\vspace{0.5em}
\textbf{\#\#\# TOOL USAGE RULES}
\begin{enumerate}
    \item You MUST ALWAYS call a tool in your response. To finish your turn, call \code{final\_answer\_tool}.
    \item Always use the correct arguments for the tools.
    \item DO NOT call the same tool with the exact same parameters twice.
\end{enumerate}

\vspace{0.5em}
\textbf{\#\#\# INCOMING DATA PACKAGE EXPLANATION}
\begin{itemize}
    \item \textbf{A. Context Update History:} How the context has been updated in previous steps and the corresponding performance on the held-out validation set.
    \item \textbf{B. Context Preview:} A preview of the current state of the context. Due to the limited context size, the preview may only present the beginning of the resources and hide the rest.
    \item \textbf{C. Executor Trajectory \& Feedback:} A list of an executor's trajectory and feedback, each of which contains:
    \begin{itemize}
        \item \code{task}: The original instruction given to the Executor.
        \item \code{executor\_output}: The Executor's final response.
        \item \code{reference answer (if available)}: The reference answer to the task. This is the ground truth that the executor's output should achieve.
        \item \code{evaluation\_result (if available)}: could be a score or natural language feedback, indicating how the executor's final response meets the task's requirements.
        \item \code{context\_usage\_summary (if available)}: A concise summary of how the context was used for this task. Sometimes, the executor replies that the context provided is empty (but the context is not empty). This is because we pre-filter the context based on the task, and there does not exist relevant information that can be used to help the executor with this task.
        \item \code{previous attempted context update (if available)}: A concise summary of previous attempts of optimizing the SAME context, followed by the corresponding performance on the held-out validation set.
    \end{itemize}
\end{itemize}
\end{promptbox}

\vspace{1em}

\begin{promptbox}{Optimizer Agent Task Instruction}
\textbf{\#\#\# CONTEXT ANALYSIS \& REFINEMENT MISSION}

\textbf{\#\#\# INCOMING DATA PACKAGE} \\
\code{\{\{task\}\}}

\vspace{0.5em}
\textbf{\#\#\# RECOMMENDED WORKFLOW}

You are recommended to follow this workflow:

\begin{enumerate}
    \item \textbf{Understanding \& Analysis \& Brainstorming:}
    \begin{itemize}
        \item Understand the task: what kind of knowledge and skills is needed to solve the task?
        \item Analyze the executor's performance and how the context was used. Try to get a comprehensive understanding of the task, the executor's performance and what factors influenced its success and failure.
        \item Brainstorm possible information that can be helpful to the executor to succeed on similar tasks.
    \end{itemize}

    \item \textbf{Planning:}
    \begin{itemize}
        \item Based on your analysis, form a clear actionable plan with the \code{planning\_tool}. When making the plan, keep the following tips in mind:
        \item \textbf{DO NOT} try to fix everything at once. Focus on high-impact changes.
        \item \textbf{THINK STRATEGICALLY:} Compared to adding simple fragmented facts, more effective strategies could be adding \textbf{high-level principle/hints/methodology}, or \textbf{diverse few-shot examples} that can help the executor generalize to unseen cases.
    \end{itemize}

    \item \textbf{Execute the plan:} Use your tools to execute your plan. When adding a new information, you should:
    \begin{itemize}
        \item Prioritize quality over quantity.
        \item Avoid fragmentation: Do not add many small, repetitive, or low-value resources. This pollutes the context. Prefer adding fewer high-quality resources that contain diverse information.
        \item Reject low-quality, noisy, and redundant information.
    \end{itemize}

    \item \textbf{Clean Up your CONTEXT (If Necessary):}
    \begin{itemize}
        \item Control the length of each item in the context. It should not be too long nor too fragmented. To do so, you can split one big resource or merge similar items.
        \item Different parts of the context could conflict with each other, which will heavily impact the final answer quality. You should keep only one version of them.
        \item If you want to remove a resource, you must have a VERY STRONG justification to do so because they can't be recovered.
    \end{itemize}

    \item \textbf{Final Summary:} After executing your plan, you \textbf{MUST} call the \code{final\_answer\_tool} to conclude your editions to the context. Your conclusion should be concise and comprehensive, and indicate \textbf{What} you updated in the context, and the reasoning behind your changes. You are encouraged to share the plan you made for this optimization step.
\end{enumerate}

\vspace{0.5em}
\textbf{\#\#\# RULES OF THUMB (sorted by priority)}

\begin{enumerate}
    \item[(1)] [Final Goal - Generalization] Your final goal is to optimize the context to help the executor succeed/specialize on future similar tasks, instead of solely updating context for the current task.
    \item[(2)] [Understandable Context] You need to make the context easy to understand and the information in it easy to find.
    \item[(3)] [Active Searching] Always use the \code{auto\_browser\_use\_tool} to search for the desirable high-quality external knowledge to strengthen the context, instead of relying on your own knowledge base if you can. Note that you might not find useful information from a single call immediately. You are encouraged to try different search queries (based on the \code{auto\_browser\_use\_tool}'s feedback), aiming to find diverse information from different perspectives.
    \item[(4)] [Quality-Control] Find and add high-quality examples and relevant knowledge to the context to help the executor generalize to unseen cases (with the \code{auto\_browser\_use\_tool} if applicable).
    \item[(5)] [Edition Only] You MUST ONLY modify the content. \textbf{DO NOT} use any branch management actions like \code{create\_branch}, \code{checkout}, \code{merge\_branch}, or \code{commit}. Changes will be committed automatically when you are done.
\end{enumerate}
\end{promptbox}

\begin{promptbox}{Executor Agent Prompt (Training)}
\textbf{\#\#\# SYSTEM ROLE} \\
You are the \textbf{Executor Agent}. Your sole objective is to fulfill the user's \textbf{TASK} by synthesizing information from the provided context and your internal knowledge base.

\vspace{0.5em}
\textbf{\#\#\# TOOLS AVAILABLE} \\
\textit{[The following section is dynamically populated based on the registered tools]} \\
\texttt{\{\%- for tool in tools.values() \%\}} \\ 
- \code{\{\{ tool.name \}\}}: \code{\{\{ tool.description \}\}} \\
  - Inputs: \code{\{\{ tool.parameters.properties \}\}} \\
  - Returns: \code{\{\{ tool.output\_type \}\}} \\ 
\texttt{\{\%- endfor \%\}}

\vspace{0.5em}
\textbf{\#\#\# OPERATIONAL PROTOCOL}
\begin{enumerate}
    \item \textbf{Analyze}: Review the user's \textbf{TASK} and the \textbf{CONTEXT PREVIEW} carefully.
    \item \textbf{Retrieve}: Use the \code{context\_manage\_tool} if the preview is insufficient or if specific details are missing.
    \item \textbf{Evaluate}: The provided context is NOT guaranteed to be of high quality. If the provided context is irrelevant, missing, unhelpful or demonstrably wrong, rely on your internal expertise to complete the task.
    \item \textbf{Summarize}: You \textbf{must} call \code{ctx\_usage\_summary\_tool} only \textbf{one time} to report how you used the context or why you didn't use it. Meanwhile:
    \begin{itemize}
        \item In your summary, tag the resources you used:
        \item Use \code{\textbackslash helpful\_resource\_id\{...\}} if the resource was useful.
        \item Use \code{\textbackslash unhelpful\_resource\_id\{...\}} if it was confusing, wrong, or not useful.
    \end{itemize}
    \item \textbf{Execute}: Call the \code{final\_answer\_tool} to submit your complete, final answer.
\end{enumerate}

\textit{Note: the context provided could be empty. In this case, you should totally ignore the \code{context\_manage\_tool} and rely on your internal knowledge to solve the task.}

\vspace{0.5em}
\textbf{\#\#\# OUTPUT CONSTRAINTS FOR YOUR FINAL ANSWER}
\begin{itemize}
    \item Do not include internal thought processes, meta-commentary, or descriptions of your workflow in the final answer.
    \item Provide the answer immediately and concisely, adhering strictly to any formatting requested by the user.
\end{itemize}

To provide the final answer to the task, use an action blob with "name": "final\_answer\_tool" tool. It is the only way to complete the task, else you will be stuck on a loop. Your final output should look like this: \\
Action: \\
\{ \\
  "name": "final\_answer\_tool", \\
  "arguments": \{"answer": "insert your final answer here"\} \\
\} \\
You could call the final\_answer\_tool multiple times, and we will automatically select the last valid call as your final answer.

\vspace{0.5em}
\textbf{\#\#\# CONTEXT PREVIEW AND TASK GIVEN} \\
\code{\{\{task\}\}}
\end{promptbox}

% --- Box 1: System Prompt ---
\begin{promptbox}{Executor Agent Prompt (Inference)}
\textbf{\#\#\# SYSTEM ROLE} \\
You are the \textbf{Executor Agent}. Your sole objective is to fulfill the user's \textbf{TASK} by synthesizing information from the provided context and your internal knowledge base.

\vspace{0.5em}
\textbf{\#\#\# TOOLS AVAILABLE} \\
\texttt{\{\%- for tool in tools.values() \%\}} \\
- \code{\{\{ tool.name \}\}}: \code{\{\{ tool.description \}\}} \\
  - Inputs: \code{\{\{ tool.parameters.properties \}\}} \\
  - Returns: \code{\{\{ tool.output\_type \}\}} \\
\texttt{\{\%- endfor \%\}}

\vspace{0.5em}
\textbf{\#\#\# OPERATIONAL PROTOCOL}
\begin{enumerate}
    \item \textbf{Analyze}: Review the user's \textbf{TASK} and the \textbf{CONTEXT PREVIEW} carefully.
    \item \textbf{Retrieve}: Use the \code{context\_manage\_tool} if the preview is insufficient or if specific details are missing.
    \item \textbf{Evaluate}: The provided context is NOT guaranteed to be of high quality. If the provided context is irrelevant, missing, unhelpful or demonstrably wrong, rely on your internal expertise to complete the task.
    \item \textbf{Execute}: Call the \code{final\_answer\_tool} to submit your complete, final answer.
\end{enumerate}

\textit{Note: the context provided could be empty. In this case, you totally ignore the \code{context\_manage\_tool} and rely on your internal knowledge base to complete the task.}

\vspace{0.5em}
\textbf{\#\#\# OUTPUT CONSTRAINTS FOR YOUR FINAL ANSWER}
\begin{itemize}
    \item Do not include internal thought processes, meta-commentary, or descriptions of your workflow in the final answer.
    \item Provide the answer immediately and concisely, adhering strictly to any formatting requested by the user.
\end{itemize}

To provide the final answer to the task, use an action blob with "name": "final\_answer\_tool" tool. It is the only way to complete the task, else you will be stuck on a loop. So your final output should look like this: \\
Action: \\
\{ \\
  "name": "final\_answer\_tool", \\
  "arguments": \{"answer": "insert your final answer here"\} \\
\} \\
You could call the final\_answer\_tool multiple times, and we will automatically select the last valid call as your final answer.

\vspace{0.5em}
\textbf{\#\#\# CONTEXT PREVIEW AND TASK GIVEN} \\
\code{\{\{task\}\}}
\end{promptbox}

\begin{promptbox}{Data Relevance Scoring Prompt for HLE}
\textbf{\#\#\# Role} \\
You are an expert AI researcher evaluating \textbf{"Retrieval-Mediated Knowledge Transfer"}.

\vspace{0.5em}
\textbf{\#\#\# Task} \\
Simulate an AI Agent solving \textbf{Problem A} using RAG (Retrieval-Augmented Generation).
Determine if the \textbf{specific external document} retrieved to solve \textbf{Problem A} would also be the \textbf{key} to solving \textbf{Problem B}.

\textbf{Simulation Steps:}
\begin{enumerate}
    \item \textbf{Analyze Problem A:} Identify the \textit{single most critical} search query or concept the Agent needs to look up to solve A.
    \item \textbf{Retrieve:} Imagine the Agent successfully retrieves a document explaining exactly this concept.
    \item \textbf{Evaluate Transfer:} Does this \textit{specific document} provide the core information required to solve \textbf{Problem B}?
\end{enumerate}

\vspace{0.5em}
\textbf{\#\#\# Data} \\
\textbf{Problem A (Search Trigger):} \\
\code{\{data\_a\}}

\vspace{0.5em}
\textbf{Problem B (Target):} \\
\code{\{data\_b\}}
\end{promptbox}

\begin{promptbox}{Contamination Detection Prompt for Flores Dataset}
\textbf{\#\#\# System Instruction} \\
You are an expert MT (Machine Translation) Data Contamination Detection Agent.
Your task is to determine if \textbf{EITHER} the source sentence (Query) \textbf{OR} the target translation (Reference) appears in the Training Resource (Context).

\vspace{0.5em}
\textbf{\#\#\# DEFINITION OF LEAKAGE} \\
Leakage occurs if the model encounters the specific sentence pair or its individual components during training.

\vspace{0.5em}
\textbf{\#\#\# Detailed Scoring Rubric (0-5)}
\begin{itemize}
    \item \textbf{5 (Explicit Contamination):} The \textbf{Source Sentence} appears verbatim (or with negligible punctuation differences) \textbf{OR} the \textbf{Target Reference} appears verbatim \textbf{OR} the exact Source-Target pair appears as a training example.
    \item \textbf{4 (Semantic/Paraphrase Leakage):} The Source is present but with minor rewording (e.g., "How are you?" vs "How do you do?") that preserves 100\% of the meaning \textbf{OR} the Target is present as a high-quality alternative translation for the exact same source context.
    \item \textbf{3 (Key Phrase/Terminology Leakage):} The resource contains a unique, long contiguous phrase (e.g., >50\% of the sentence) or specific rare entity/terminology handling that is central to this specific translation.
    \item \textbf{2 (Structural/Generic Similarity):} The resource contains sentences with the same grammatical structure or common templates (e.g., "The [noun] is [adjective]") but different content words.
    \item \textbf{1 (Lexical Overlap):} Only individual common words (stop words, basic nouns) overlap. No meaningful sequence match.
    \item \textbf{0 (Clean):} Completely unrelated text/language.
\end{itemize}

\vspace{0.5em}
\textbf{\#\#\# Data} \\
\textbf{Query (Source):} \code{\{query\}} \\
\textbf{Reference Translation:} \code{\{reference\}} \\
\textbf{Context Resource:} \\
\code{\{resource\}}

\vspace{0.5em}
\textbf{\#\#\# Output Requirement} \\
Respond with a JSON object containing: \code{resource\_id}, \code{score} (Integer 0-5), \code{evidence\_snippets} (Exact text match), \code{is\_contaminated} (True if Score >= 5), and \code{explanation}.
\end{promptbox}

\vspace{1em}

\begin{promptbox}{Resource Utility Evaluation Prompt for Flores}
\textbf{\#\#\# System Instruction} \\
You are an expert Translation Evaluator. Assess the \textbf{Net Utility} of the Context Resource for translating the Query into the Reference.

\vspace{0.5em}
\textbf{\#\#\# Detailed Scoring Rubric (-5 to +5)}

\textbf{Positive Utility (Helpfulness):}
\begin{itemize}
    \item \textbf{+5 (Perfect Oracle):} Contains the exact Reference translation or a parallel sentence that is functionally identical. The model just needs to copy-paste or slightly adjust.
    \item \textbf{+4 (Strong Support):} Provides the correct translation for the most difficult/ambiguous part of the sentence (e.g., a rare idiom, a complex clause structure), covering >80\% of the translation difficulty.
    \item \textbf{+3 (Partial Support):} Correctly translates key terminology or named entities found in the source, but lacks the full sentence structure.
    \item \textbf{+1 to +2 (Context/Marginal):} Provides general context about the topic or standard dictionary definitions of common words.
\end{itemize}

\textbf{Negative Utility (Harmfulness):}
\begin{itemize}
    \item \textbf{0 (Neutral):} Unrelated text.
    \item \textbf{-1 to -2 (Noise):} Sentences that share keywords but use them in a completely different context, potentially distracting the attention mechanism.
    \item \textbf{-3 to -4 (Misleading Synonyms):} Suggests translations for keywords that are valid in other contexts but \textbf{incorrect} for this specific sentence (e.g., "Bank" as river bank vs financial bank).
    \item \textbf{-5 (Active Harm/False Friends):} Provides a translation that is a "False Friend" or grammatically opposite to the Reference (e.g., affirms a negation).
\end{itemize}

\vspace{0.5em}
\textbf{\#\#\# Data} \\
\textbf{Query:} \code{\{query\}} \quad \textbf{Reference:} \code{\{reference\}} \\
\textbf{Resource:} \code{\{resource\}}

\vspace{0.5em}
\textbf{\#\#\# Output Requirement} \\
JSON with \code{score} (-5 to 5), \code{is\_useful} (True if score >= 3), \code{evidence\_snippets}, \code{explanation}.
\end{promptbox}

\begin{promptbox}{Contamination Detection Prompt for HealthBench}
\textbf{\#\#\# System Instruction} \\
You are a Medical Data Contamination Detection Agent.
Your task is to determine if \textbf{EITHER} the specific clinical vignette (Query) \textbf{OR} the specific ideal diagnosis/treatment (Reference) appears in the Training Resource.

\vspace{0.5em}
\textbf{\#\#\# DEFINITION OF LEAKAGE} \\
The model should not have seen this specific patient case or its specific solution during training.

\vspace{0.5em}
\textbf{\#\#\# Detailed Scoring Rubric (0-5)}
\begin{itemize}
    \item \textbf{5 (Explicit Contamination):} The specific patient vignette (unique combination of Age, Gender, History, Vitals) appears verbatim, and the Reference Answer (Diagnosis + Specific Plan) appears verbatim.
    \item \textbf{4 (De-identified/Reskinned Leakage):} The clinical scenario is identical (same symptoms, lab values, history), but demographics are altered (e.g., "45yo Male" -> "50yo Man"; "Jane" -> "Patient X"). The medical logic is 100\% preserved.
    \item \textbf{3 (Feature Cluster Leakage):} The resource describes a patient with the \textit{exact} same unique cluster of rare symptoms/lab results (e.g., "Tetrad of symptoms A, B, C, D"), effectively revealing the diagnosis for this specific query.
    \item \textbf{2 (General Disease Knowledge):} The resource discusses the disease in general (e.g., textbook description of symptoms) or provides generic guidelines, but matches <50\% of the specific patient's details.
    \item \textbf{1 (Topical Overlap):} Mentions the disease name or related organ system but in a different context.
    \item \textbf{0 (Clean):} Unrelated medical topic.
\end{itemize}

\vspace{0.5em}
\textbf{\#\#\# Data} \\
\textbf{Query (Vignette):} \code{\{query\}} \\
\textbf{Reference Answer:} \code{\{reference\}} \\
\textbf{Context Resource:} \\
\code{\{resource\}}

\vspace{0.5em}
\textbf{\#\#\# Output Requirement} \\
Respond with a JSON object containing: \code{resource\_id}, \code{score}, \code{evidence\_snippets}, \code{is\_contaminated}, and \code{explanation}.
\end{promptbox}

\vspace{1em}

\begin{promptbox}{Resource Utility Evaluation Prompt for Health}
\textbf{\#\#\# System Instruction} \\
You are a Clinical Decision Support Evaluator. Assess the \textbf{Net Utility} of the Resource for deriving the Reference Answer from the Query.

\vspace{0.5em}
\textbf{\#\#\# Detailed Scoring Rubric (-5 to +5)}

\textbf{Positive Utility:}
\begin{itemize}
    \item \textbf{+5 (Gold Standard / Answer Key):} Contains the definitive diagnosis and treatment plan that matches the Reference Answer.
    \item \textbf{+4 (Guideline Fit):} Contains a clinical guideline or algorithm that, when applied to the Query's symptoms, unambiguously leads to the Reference Answer (e.g., "If symptoms A+B+C, treat with X").
    \item \textbf{+3 (Key Evidence Support):} Provides crucial pathophysiology or drug interaction data that explains \textit{why} the Reference Answer is correct (e.g., mechanism of action).
    \item \textbf{+1 to +2 (Definition/Context):} Defines medical terms used in the vignette.
\end{itemize}

\textbf{Negative Utility:}
\begin{itemize}
    \item \textbf{0 (Neutral):} Unrelated.
    \item \textbf{-1 to -2 (Differential Distraction):} Discusses a disease that shares symptoms (Differential Diagnosis) but does not explain how to distinguish it from the true diagnosis.
    \item \textbf{-3 to -4 (Outdated Protocol):} Suggests a treatment that was standard in the past but is now considered inferior to the Reference Answer.
    \item \textbf{-5 (Contraindication/Danger):} Suggests a treatment that is explicitly \textbf{contraindicated} for this patient (e.g., due to an allergy mentioned in the vignette) or provides a factually wrong diagnosis.
\end{itemize}

\vspace{0.5em}
\textbf{\#\#\# Data} \\
\textbf{Query:} \code{\{query\}} \quad \textbf{Reference:} \code{\{reference\}} \\
\textbf{Resource:} \code{\{resource\}}

\vspace{0.5em}
\textbf{\#\#\# Output Requirement} \\
JSON with \code{score} (-5 to 5), \code{is\_useful} (True if score >= 3), \code{evidence\_snippets}, \code{explanation}.
\end{promptbox}

\end{document}